\documentclass[10pt,twocolumn,letterpaper]{article}

\pdfoutput=1

\usepackage[pagenumbers]{cvpr} 

\usepackage[dvipsnames]{xcolor}
\usepackage{graphicx}
\usepackage{amsmath}
\usepackage{amssymb}
\usepackage{booktabs}
\usepackage{float}
\usepackage{enumerate}

\usepackage{mathtools} 
\usepackage{booktabs} 
\usepackage{tikz} 
\usepackage{amsfonts}
\usepackage{multirow}
\usepackage{multicol}
\usepackage{color,colortbl}
\usepackage{xspace}
\usepackage{wrapfig}
\usepackage{subcaption}
\usepackage{bbm}
\usepackage{bm}
\usepackage{url}

\def\eg{e.g.,\ }               
\def\ie{i.e.,\ }               
\def\vs{vs.\ }                 

\DeclareMathOperator*{\argmax}{\arg\!\max}

\definecolor{LavenderBlue}{rgb}{0.7020,    0.8039,    0.8902}
\definecolor{Lightapricot}{rgb}{0.9961,    0.8510,    0.6510}
\definecolor{thirdtablecolor}{rgb}{0.8706,    0.7961,    0.8941}

\newcommand{\heading}[1]{\noindent\textbf{#1}}

\long\def\ignorethis#1{}

\definecolor{demphcolor}{RGB}{100,100,100}

\newlength\pagetopmargin
\newlength\figcapmargin
\newlength\figmargin
\newlength\tablecapmargin
\newlength\tablemargin

\usepackage{soul}

\newcommand{\algoNameFull}{EventCLIP\xspace}

\definecolor{cvprblue}{rgb}{0.21,0.49,0.74}
\usepackage[pagebackref,breaklinks,colorlinks,citecolor=cvprblue]{hyperref}

\usepackage[capitalize]{cleveref}
\crefname{section}{Sec.}{Secs.}
\Crefname{section}{Section}{Sections}
\Crefname{table}{Table}{Tables}
\crefname{table}{Tab.}{Tabs.}
\Crefname{figure}{Figure}{Figures}
\crefname{figure}{Fig.}{Figs.}

\def\paperID{8571} 
\def\confName{CVPR}
\def\confYear{2024}

\title{\algoNameFull: Adapting CLIP for Event-based Object Recognition}

\author{
Ziyi Wu$^{1,2}$, Xudong Liu$^1$, Igor Gilitschenski$^{1,2}$ \\
$^1$University of Toronto, $^2$Vector Institute \\
{\tt\small \{ziyiwu,xudong,gilitschenski\}@cs.toronto.edu}
}

\begin{document}

\maketitle

\begin{abstract}

Recent advances in zero-shot and few-shot classification heavily rely on the success of pre-trained vision-language models (VLMs) such as CLIP.
Due to a shortage of large-scale datasets, training such models for event camera data remains infeasible.
Thus, adapting existing VLMs across modalities to event vision is an important research challenge.
In this work, we introduce \algoNameFull, a novel approach that utilizes CLIP for zero-shot and few-shot event-based object recognition.
We first generalize CLIP's image encoder to event data by converting raw events to 2D grid-based representations.
To further enhance performance, we propose a feature adapter to aggregate temporal information over event frames and refine text embeddings to better align with the visual inputs.
We evaluate \algoNameFull on N-Caltech, N-Cars, and N-ImageNet datasets, achieving state-of-the-art few-shot performance.
When fine-tuned on the entire dataset, our method outperforms all existing event classifiers.
Moreover, we explore practical applications of \algoNameFull including robust event classification and label-free event recognition, where our approach surpasses previous baselines designed specifically for these tasks.

\end{abstract}


\section{Introduction}

Event-based cameras have recently gained significant interest in computer vision due to their high temporal resolution, low energy consumption, and high dynamic range properties~\citep{EventVisionSurvey}.
Event-based vision has shown promising results in various applications, such as object recognition~\citep{EST,DiST-N-IN,HATS-N-Cars}, detection~\citep{1MpxDet,RVT,ASTMNet}, tracking~\citep{EventTracking1,EventTracking3,EKLT}, and optical flow estimation~\citep{EV-FlowNet,E-RAFT,EventOptFlow1}.
However, this novel imaging modality poses unique challenges, including the need for specialized models to handle the asynchronous nature of events, and the lack of large-scale datasets.
As in classical recognition problems, newly captured event data can contain objects from categories that are not present in the training set of deployed models.
In such cases, trained models will fail, and it may be infeasible to re-train the model every time a new object category is introduced, motivating the need for event-based zero-shot and few-shot recognition systems.

\begin{figure}[t]
    \vspace{\pagetopmargin}
    \centering
    \hspace{-1mm}
    \begin{subfigure}{1.02\linewidth}
        \includegraphics[width=1.0\linewidth]{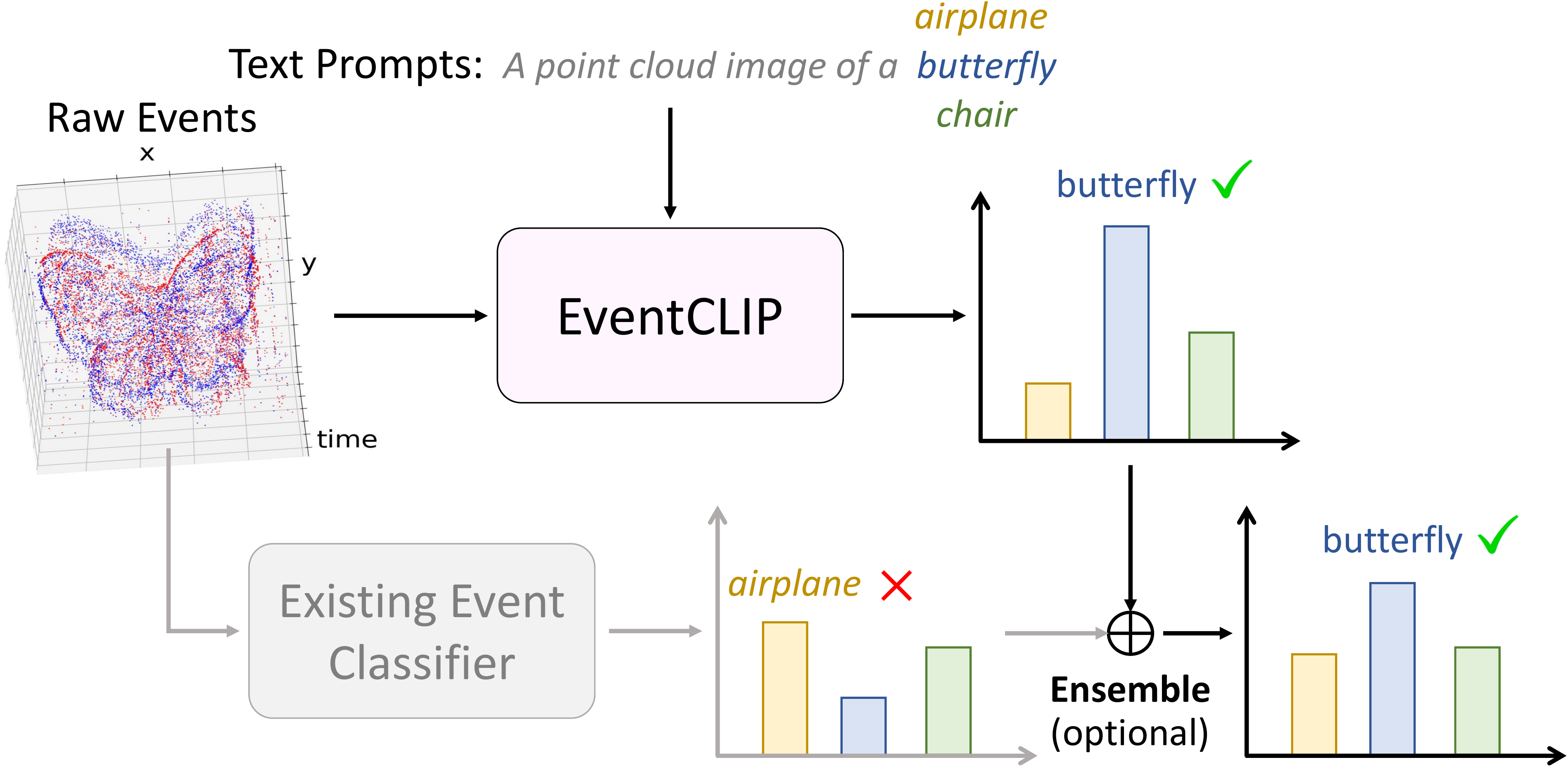}
        \vspace{-5mm}
        \caption{Open-set and robust classification}
    \end{subfigure}
    \\
    \vspace{-2mm}
    \hspace{-1mm}
    \begin{subfigure}{1.02\linewidth}
        \includegraphics[width=1.0\linewidth]{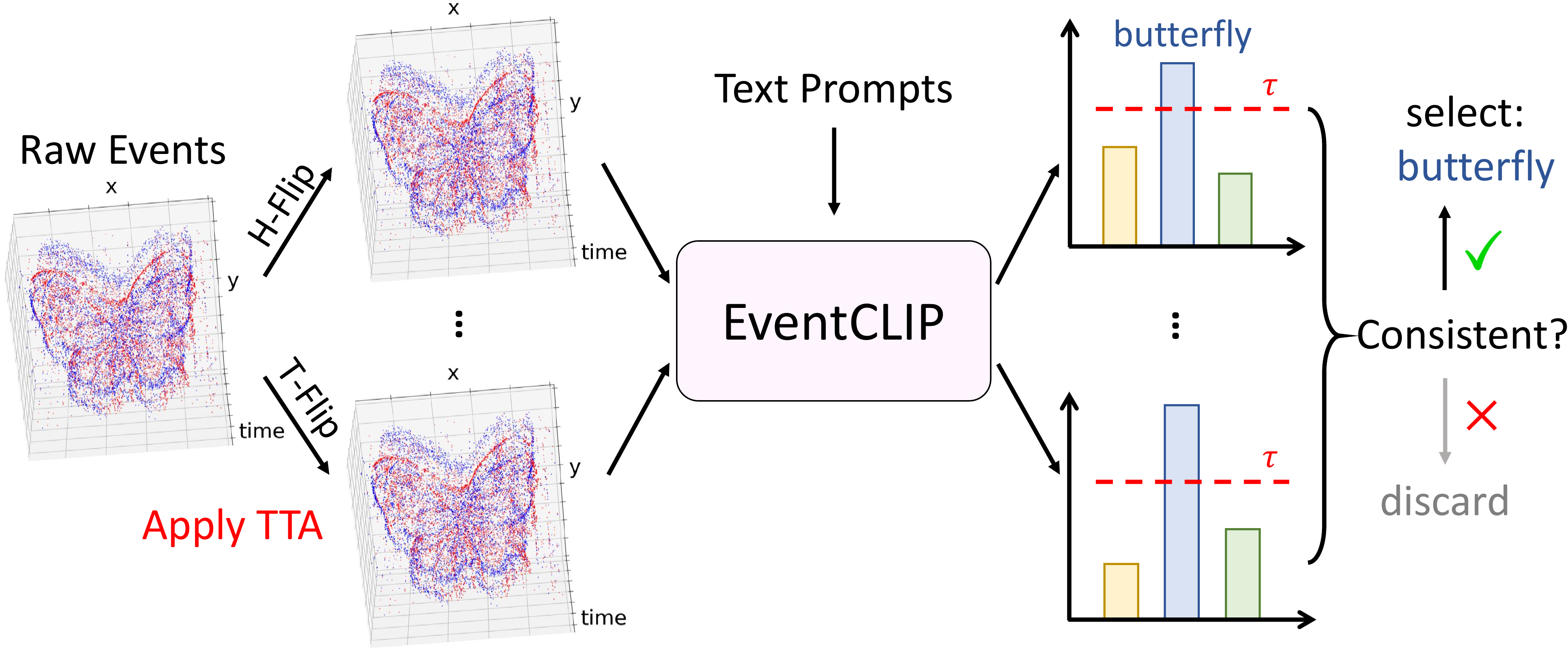}
        \vspace{-4mm}
        \caption{Learning from unlabeled data}
    \end{subfigure}
    \vspace{\figcapmargin}
    \vspace{-4.5mm}
    \caption{
    Existing event-based classifiers are trained from scratch on closed-set datasets.
    They fail on events with unseen categories, camera motions, or lighting conditions.
    Instead, our method utilizes pre-trained CLIP for zero-shot or few-shot open-world event recognition.
    Moreover, \algoNameFull can be applied to the task of (a) robust event classification by ensembling with trained models, and (b) label-free learning by training on generated pseudo labels.
    }
    \label{fig:teaser}
    \vspace{\figmargin}
    \vspace{-1mm}
\end{figure}

In frame-based vision, pre-trained vision-language models (VLMs) such as CLIP~\citep{CLIP} have shown remarkable success in zero-shot and few-shot learning tasks.
Trained on large-scale datasets, these models try to map paired images and texts to an aligned feature space.
Open-world zero-shot classification is made possible by leveraging the feature similarity between unseen objects and texts with novel category names~\citep{CLIP,ALIGN}.
Recently, several works have designed data-efficient methods to adapt CLIP under the few-shot learning setting for better accuracy~\citep{CLIP-Adapter,CoOp,CoCoOp,Tip-Adpater,WiSE-FT}.
However, there is currently no large-scale event-text dataset available, making it impossible to train such event-language models from scratch.
This motivates us to ask the question: can 2D pre-trained VLMs be transferred to event-based vision and realize zero-shot or few-shot object recognition?

In this work, we propose \algoNameFull as the first attempt introducing CLIP to event-based visual understanding.
To bridge the gap between asynchronous event data and CLIP's frame-based input representation, we split an event stream into multiple time windows, and convert each of them into a 2D frame.
Following \citet{CLIP}, text prompts are constructed by placing class names into hand-crafted templates, and text features are extracted as the zero-shot classifier weight.
Each event frame is classified by CLIP individually, and the final result is obtained by simple voting.

Although \algoNameFull can achieve zero-shot recognition, its performance still lags behind existing classifiers trained on event domain data.
We thus propose to learn lightweight adapters to refine the pre-trained CLIP features.
Different from previous work which only adapts one image feature~\citep{CLIP-Adapter} or a fixed number of features in pre-defined orders~\citep{PointCLIP}, the number and order of our event features depend on the camera trajectory.
We thus design a Transformer-based adapter to aggregate temporal information from multiple frames.
We also fine-tune text features as the weight of the output fully-connected layer in a classifier.
With these designs, \algoNameFull achieves more data-efficient few-shot learning compared to existing event-based classifiers.

Finally, we explore more applications of \algoNameFull.
We discover that the 2D pre-trained knowledge in CLIP, acquired from Internet-scale data, synergizes with the domain knowledge in models trained purely on event data.
Therefore, we directly ensemble \algoNameFull and existing event classifiers, which consistently improves their accuracy by more than 10\% on N-ImageNet robustness subsets featuring unseen camera motions and lightnings~\cite{DiST-N-IN}.
In addition, since \algoNameFull can classify unseen objects, we use it to label raw events. 
Leveraging the spatio-temporal property of events, we are able to select high-quality pseudo labels and achieve state-of-the-art unsupervised classification results.

In summary, this work makes four main contributions:
\textbf{(i)} the first zero-shot open-world event-based object recognition method using CLIP,
\textbf{(ii)} a Transformer-based feature adapter tailored to event temporal information aggregation,
\textbf{(iii)} state-of-the-art few-shot and fine-tuning results on three datasets, and
\textbf{(iv)} significantly improved robustness and unsupervised classification accuracy on N-ImageNet subsets.

\section{Related Work}\label{sec:related-work}

We briefly review recent works in event-based recognition, bridging general and event-based vision, and CLIP-based transfer learning, which is further expanded in Appendix~\ref{app:more-related-work}.

\heading{Deep Learning for Event-based Classification.}
Depending on the utilization of the asynchronous nature of events, existing event-based classifiers can be mainly categorized into two classes, namely, synchronous and asynchronous methods.
Synchronous models aggregate events to a grid-based representation, followed by standard modules such as Convolutional Neural Networks (CNNs)~\citep{HATS-N-Cars,EST,Video2Event,DiST-N-IN,EventGraftNet}.
Significant efforts have been made to achieve efficient and expressive event-to-frame conversion, such as binarized event occurrence~\citep{BinEvImg}, event counts~\citep{EH}, and sorted event timestamps~\citep{STS}.
Recently, EST~\citep{EST} has achieved state-of-the-art results with an end-to-end learnable event-to-frame conversion pipeline.
To improve the robustness against data noise, DiST~\citep{DiST-N-IN} proposes to suppress noisy events with their spatio-temporal relationships, which has proved effective under camera motion and lighting change in data capture.
Compared to asynchronous methods~\cite{SlideGCN,AEGNN,EventSparseConv}, synchronous models achieve consistently better results across datasets.
As our primary goal is to achieve high accuracy instead of efficiency, we adopt representative synchronous event-based classifiers as our baselines in the experiments.

\heading{Bridging General and Event-based Vision.}
Inspired by the great success of classical computer vision, several works have introduced techniques from frame-based vision to process event data.
Some papers focus on reconstructing natural images from events, and then apply conventional deep models on the converted frames~\citep{E2VID,E2VID-PAMI,BetterE2VID,FireNet,E2VID-Transformer,Ev-LaFOR}.
Yet, they often introduce large computational overhead which is at odds with event cameras' low-latency nature.
Closer to ours are methods that transfer knowledge learned from images to event-based models~\citep{EventGraftNet,Evdistill,ESS_EvSegFromImg,WormholeLearningRSS,WormholeLearningICRA,EventDA}.
However, they either require paired recordings of images and events, or massive labels on image data.
In this work, we utilize CLIP pre-trained on RGB image-text pairs for data-efficient event-based classification.
Our method converts events into frames via simple counting, and directly applies CLIP for zero-shot classification.
We can further boost its performance via few-shot feature adaptation, without the need for paired RGB images or large amounts of labels.

\heading{CLIP-based Few-Shot Transfer Learning.}
Transfer learning aims to leverage large pre-trained models to facilitate learning in data-scarce scenarios.
In event-based object recognition, existing methods also utilize models pre-trained on RGB images from ImageNet~\citep{ImageNet} as their backbones to improve performance~\citep{EH,EST,DiST-N-IN,STS}.
Trained on millions of image-text pairs, CLIP~\citep{CLIP} learns transferable representations that are useful for downstream tasks.
To further enhance its accuracy, some methods~\cite{CoOp,CoCoOp,KgCoOp} insert learnable text tokens to perform task-specific prompt tuning, which requires backpropagation over the heavy text encoder.
CLIP-Adapter~\cite{CLIP-Adapter}, Tip-Adapter~\cite{Tip-Adpater}, and WiSE-FT~\cite{WiSE-FT} instead learn lightweight CLIP feature adapters.

Besides 2D image classification, CLIP has also been extended to 2D detection~\cite{ViLD,Detic,MDETR}, segmentation~\cite{DenseCLIP1,DenseCLIP2}, and video analysis~\cite{ActionCLIP,VideoCLIPPrompting}.
Our work is inspired by PointCLIP~\citep{PointCLIP,PointCLIP_V2}, which projects point clouds to multi-view images for CLIP-based zero-shot and few-shot 3D shape recognition.
Unlike PointCLIP, event data usually only capture the boundary information of objects, which poses a greater domain gap than point clouds as point clouds often depict complete object surfaces.
In addition, we design a Transformer-based adapter for event temporal information fusion, while PointCLIP simply uses an MLP since their multi-view projections follow a pre-defined order.

\section{Method}\label{sec:method}

\begin{figure*}[t]
    \vspace{\pagetopmargin}
    \vspace{-1mm}
    \centering
    \includegraphics[width=0.99\linewidth]{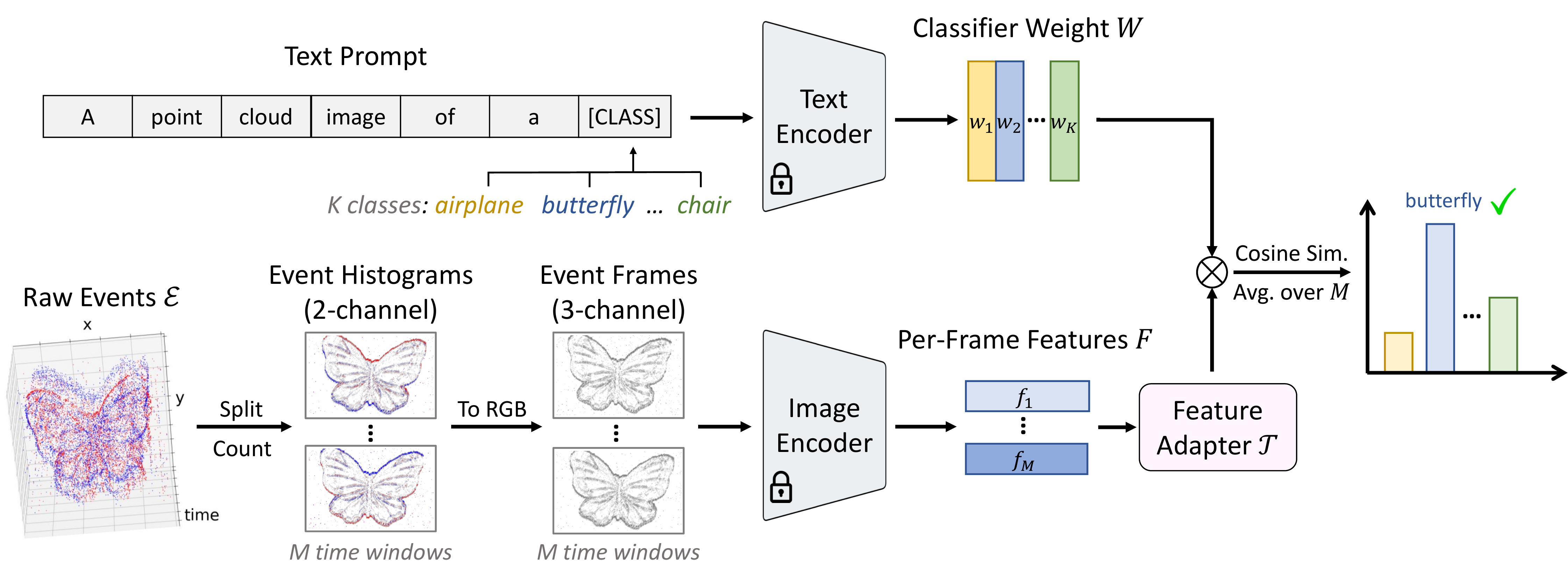}
    \vspace{\figcapmargin}
    \caption{
    \textbf{\algoNameFull overview}.
    Given raw events $\mathcal{E}$, we split it to $M$ time windows and construct $M$ 2-channel event histograms.
    They are converted to 3-channel RGB frames with a color map, and fed to CLIP to obtain image features $F$.
    Meanwhile, we construct text prompts with $K$ class names, and utilize CLIP to extract text features as the classifier weight $W$.
    $F$ is updated with a feature adapter $\mathcal{T}$ to fuse temporal information, and its cosine similarity with $W$ serves as the classification output.
    $\mathcal{T}$ is an identity function under the zero-shot setting, and a Transformer encoder in few-shot learning.
    The final result is obtained by averaging predictions from $M$ frames.
    }
    \label{fig:pipeline}
    \vspace{\figmargin}
\end{figure*}

\algoNameFull builds upon large-scale pre-trained CLIP~\cite{CLIP} (\cref{sec:revisit-clip}) and converts event streams to 2D images for zero-shot event understanding (\cref{sec:zero-shot-eventclip}).
When a few labeled examples are provided, we learn lightweight feature adapters to further improve the few-shot accuracy (\cref{sec:few-shot-eventclip}).
Finally, we extend \algoNameFull to more tasks including robust event classification and label-free event recognition in \cref{sec:apply-eventclip}.
The overall pipeline of \algoNameFull is illustrated in \cref{fig:pipeline}.

\subsection{Background: CLIP-based Image Classification}\label{sec:revisit-clip}

The training objective of CLIP is to map images and texts to a joint embedding space.
CLIP consists of two encoders for image and text inputs, respectively.
During training, given a batch of image-text pairs, CLIP maximizes the cosine similarity between embeddings of positive pairs, while minimizing it for negative pairs using a contrastive loss.
CLIP is trained on a collection of 400 million web-crawled image-text data.
The large pre-training dataset enables the incorporation of diverse visual concepts, thereby enhancing the transferability of the learned features to downstream tasks.

As CLIP is trained to match image and text features, it naturally lends itself to zero-shot classification.
Formally, let $\bm{f}_{\bm{x}}$ be the image feature extracted by CLIP's image encoder for an image $\bm{x}$.
Meanwhile, we construct text inputs by placing category names into a pre-defined template such as ``\texttt{a photo of a [CLASS]}", and leverage CLIP's text encoder to extract a set of $K$ features $W = \{\bm{w}_i\}_{i=1}^K$, where $K$ denotes the number of classes.
The probability of predicting class $i$ given image $\bm{x}$ is then computed as:
\begin{equation}\label{eq:clip-zero-shot-cls}
    p(y = i | \bm{x}) = \frac{\exp(\cos(\bm{w}_i, \bm{f}_{\bm{x}})/ \tau) }{\sum_{j=1}^K \exp(\cos(\bm{w}_j, \bm{f}_{\bm{x}})/ \tau)},
\end{equation}
where $\cos(\cdot, \cdot)$ denotes the cosine similarity between two vectors, and $\tau$ is a scaling factor learned by CLIP.
The zero-shot inference does not require any in-domain training data to fine-tune the model, but can achieve competitive results with fully supervised baselines on 2D image datasets~\citep{CLIP}.

\subsection{Zero-Shot Event Understanding}\label{sec:zero-shot-eventclip}

Prior work in event recognition has shown that raw events can be converted to meaningful 2D frames such as edge maps~\citep{EventVisionSurvey}.
This motivates us to prompt VLMs pre-trained in the RGB domain to process event camera data.
We adopt CLIP due to its public availability and importance in the literature.
However, the presented approach also generalizes to other VLMs as we will show in the experiments.

\heading{Bridging the Modality Gap.}
Event cameras record brightness changes at each sensing pixel, and output a sequence of events $\mathcal{E} = \{e_i = (x_i, y_i, t_i, p_i)\}$, each parameterized by its spatial position $(x_i, y_i)$, triggered timestamp $t_i$, and polarity $p_i \in \{-1, 1\}$.
Since raw events are asynchronous and sparse, they are represented as a set, which differs from the grid-like representations CLIP requires.
To bridge this modality gap, we convert raw events into 2D frames.
Specifically, we split an event stream $\mathcal{E}$ to $M$ time windows $\{\mathcal{E}_i\}_{i=1}^M$ by grouping every $N$ consecutive events.
Grouping with event count ensures better robustness against camera speed compared to using fixed time intervals~\citep{DiST-N-IN}.
For each $\mathcal{E}_i$, we construct a 2-channel histogram by counting the number of positive and negative events at each pixel.
To obtain a 3-channel image, we first normalize the histogram to the range of $[0, 1]$, and then colorize it with a pre-defined RGB color map.
Finally, we set the empty pixels to pure white for better visual quality following \citet{EventMAE}.


\heading{Zero-Shot Classification.}
After converting raw events to $M$ 2D frames, we leverage CLIP's image encoder to extract their features $F = \{\bm{f}_i\}_{i=1}^M$.
Following \citet{CLIP}, we then construct text prompts with class names and hand-crafted templates, and use CLIP to extract text embeddings $W = \{\bm{w}_i\}_{i=1}^K$.
The zero-shot prediction for each time window can be computed with Eq.~\ref{eq:clip-zero-shot-cls}.
Here, the text template should reflect the domain-specific knowledge about event data, such as the visual property of the converted frames.

To obtain the final classification output, we need to aggregate predictions from $M$ time windows.
PointCLIP~\citep{PointCLIP,PointCLIP_V2} also faces this problem as they project a point cloud to multiple views. 
They simply assign hyper-parameters to weigh the importance of each view, which are fixed through the dataset.
This is feasible since the 3D point clouds they considered are all aligned to a canonical pose, and thus the projected views for different data follow the same order (\eg \{front, right, back, left, top, bottom\}).
However, the temporal order of our time windows depends on the event camera's trajectory, which varies a lot across data samples, making a pre-defined set of weights sub-optimal.
Inspired by DeepSets~\citep{DeepSets}, we select the order-invariant set operation mean-pooling to average the classification probabilities from all time windows as the final prediction output.

\subsection{Few-Shot Feature Adaptation}\label{sec:few-shot-eventclip}

With the event-to-frame conversion pipeline, we successfully transform a 2D CLIP into a zero-shot event classifier for the ``unseen" event camera data.
However, zero-shot \algoNameFull still underperforms domain-specifically trained classifiers.
To close the accuracy gap, we consider the few-shot setting, where a few labeled examples are available per class.
With limited data, it is impossible to fine-tune the entire model which will lead to severe overfitting.
Instead, we only refine the features extracted by a frozen CLIP model.

\heading{Image Feature Adapter.}
Our goal is to incorporate the event domain knowledge into the extracted image features $\{\bm{f}_i\}_{i=1}^M$ to obtain a refined representation $\{\bm{f}^*_i\}_{i=1}^M$.
Prior works simply apply an MLP to update features since there is either only one feature vector per sample (2D image)~\citep{CLIP-Adapter,WiSE-FT}, or the visual features follow a fixed order (3D point cloud)~\citep{PointCLIP,PointCLIP_V2}.
For example, PointCLIP concatenates the multi-view features into one vector and feeds it to the MLP-based adapter.
Their performance will degrade significantly if we shuffle the order of projection views.
In contrast, as discussed above, the final prediction of \algoNameFull should be \textit{order-invariant}.
Therefore, the output of our image feature adapter $\{\bm{f}^*_i\}_{i=1}^M$ should be \textit{permutation-equivariant} to the input features $\{\bm{f}_i\}_{i=1}^M$.
In addition, we need an architecture that can process an arbitrary number of inputs as the number of time windows $M$ varies across samples.
Inspired by a recent CLIP-based video classifier~\citep{VideoCLIPPrompting}, we apply a lightweight 2-layer Transformer encoder~\citep{Attention}, $\mathcal{T}$, to aggregate the temporal information of event streams.
To avoid overfitting, we employ residual connections from CLIP features to the Transformer output features $\{\tilde{\bm{f}}_i\}_{i=1}^M$:
\begin{equation}\label{eq:visual-adapter}
    \{\tilde{\bm{f}}_i\}_{i=1}^M = \mathcal{T}(\{\bm{f}_i\}_{i=1}^M),\ \ \bm{f}^*_i = \alpha \bm{f}_i + (1 - \alpha) \tilde{\bm{f}}_i,
\end{equation}
where $\alpha$ is a hyper-parameter controlling the portion of the original CLIP knowledge.
After applying the visual adapter, we use the updated image features $\{\bm{f}^*_i\}_{i=1}^M$ and the text features $W$ to perform classification as done in the zero-shot setting.
Thanks to the order invariance property, our few-shot \algoNameFull is also more robust against different camera motions during the data capture process.

\heading{Text Feature Adapter.}
Recently, several works have studied data-efficient fine-tuning of CLIP's text branch~\citep{CoOp,CoCoOp}.
As pointed out by~\citet{SyntheticData4CLS}, all these methods aim at learning a better classifier weight $W = \{\bm{w}_i\}_{i=1}^K$.
Therefore, we follow them to adopt the simple Classifier Tuning method~\citep{WiSE-FT}, by fine-tuning $W$ with gradient descent.
In our initial experiments, Classifier Tuning indeed achieves competitive performance with more complicated tuning methods~\cite{CoOp,KgCoOp}, while requiring much less computation.

\subsection{Extensions of \algoNameFull}\label{sec:apply-eventclip}


\heading{Robust classification.}
Existing event-based classifiers are trained from scratch on event data.
These datasets are often captured under limited environment variations~\citep{DiST-N-IN}.
Thus, the model performance degrades drastically when tested on unseen settings, such as changes in lighting or camera motion.
On the contrary, CLIP is trained on Internet-scale data, thus exhibiting high robustness against data corruption.
A natural idea is thus to ensemble the two models for joint prediction.
Specifically, we simply average the predicted logits from a pre-trained event-based classifier and a zero-shot or few-shot \algoNameFull as the final output.
As we will show in the experiments, the domain-specific event knowledge and the 2D pre-trained knowledge are able to complement each other, leading to state-of-the-art model robustness.

\heading{Label-free learning.}
In many practical scenarios, we have access to not only a few labeled data but also massive unlabeled events.
The extreme setting in this line is label-free learning, where we only have raw events without any labels.
In both cases, we can leverage \algoNameFull to create pseudo labels, and then fine-tune the model on them.
To generate reliable pseudo labels, we run predictions on multiple augmented versions of the event, and only select data with a consistent predicted label.
Formally, based on the fact that an event stream should remain the same class after horizontal flip and temporal reverse, given an event, we create four versions of it by applying the augmentations combinatorially. 
Then, we discard events with an inconsistent predicted class.
To further improve the label quality, we employ a threshold $\tau$ to select high-confident samples, and only take the top-$k$ predictions per class for balanced model training.
See Appendix~\ref{app:implement-details} for implementation details in this setting.

\section{Experiments}\label{sec:exp}

In \cref{sec:zero-shot-cls}, we study the best design choices to transfer CLIP's pre-trained knowledge to event camera data.
Then, we show the performance gain from limited training data in few-shot learning (\cref{sec:few-shot-cls}).
When more data are available, \algoNameFull can achieve state-of-the-art accuracy by fine-tuning the entire model (\cref{sec:fine-tune}).
In \cref{sec:robust-cls}, we leverage our method to improve the robustness of existing event classifiers by ensemble.
Finally, in \cref{sec:unsup-learning}, we demonstrate unsupervised learning from raw events with our method.

\subsection{Experimental Setup}\label{sec:exp-setup}

\heading{Datasets.}
We use three public datasets in our experiments: N-Caltech~\citep{N-Caltech}, N-Cars~\citep{HATS-N-Cars}, and N-ImageNet~\citep{DiST-N-IN}.
\textit{N-Caltech} contains 8,246 samples from 101 classes, recorded by a moving 180 $\times$ 240 resolution ATIS system~\citep{ATIS} in front of a monitor displaying still images from the original RGB Caltech101 dataset~\citep{Caltech101}.
In contrast, \textit{N-Cars} provides event streams recorded by the ATIS system in a real-world urban environment.
It contains 12,336 samples of the class car and 11,693 samples of the class background.
Similar to N-Caltech, \textit{N-ImageNet} is the event camera version of ImageNet~\citep{ImageNet}.
As the largest event camera dataset, it contains 1.78 million event streams and 1,000 classes.
The data were captured with a moving 480 $\times$ 640 resolution Samsung DVS Gen3 event camera~\citep{DVE_Gen3}.
N-ImageNet also provides variants of the test set captured with different camera motions and brightness, serving as a benchmark to evaluate the robustness of event classifiers.
See Appendix~\ref{app:NIN-variants} for detailed descriptions of each variant.
For few-shot training, we randomly sample a subset of data from each category.
We always report the results on the entire test set.

\heading{Baselines.}
We compare \algoNameFull with current state-of-the-art event-based classifiers, namely, EST~\citep{EST}, Event Histogram~\citep{EH}, Sorted Time Surface~\citep{STS}, and DiST~\citep{DiST-N-IN}.
See Appendix~\ref{app:implement-details} for their implementation details.
Notice that, we use ResNet34~\citep{ResNet} pre-trained on the RGB ImageNet~\citep{ImageNet} as the backbone for all the baselines following their original paper.
For DiST and EST, we also tested larger backbones such as ResNet101 and ViT-L~\citep{ViT}, but did not observe clear improvement as will be shown later.
We will introduce other baselines in each task below.

\heading{Our Implementation Details.}
To convert event streams into frames, we set the number of events per time window $N$ as 20,000, 10,000, and 70,000 on N-Caltech, N-Cars, and N-ImageNet, respectively.
This accounts for each dataset's event camera resolution.
For colorizing the events, \ie converting the 2-channel event histograms to 3-channel RGB images, we simply use a gray-scale color map by multiplying both positive and negative event counts with [127, 127, 127].
For the pre-trained CLIP, we adopt the variant with the ViT-L/14~\cite{ViT} image encoder.
We select ``\texttt{a point cloud image of a [CLASS]}" as the text template.

\begin{table}[t]
    \vspace{-1mm}
    \centering
    \small
    \setlength{\tabcolsep}{6pt}
    \begin{tabular}{cccc}
        \toprule
        \textbf{Dataset} & N-Caltech & N-Cars & N-ImageNet \\
        \midrule
        Acc. & 69.67 & 82.28 & 20.78 \\
        \bottomrule
    \end{tabular}
    \vspace{\tablecapmargin}
    \caption{
    \textbf{Zero-shot classification accuracy} (\%) of \algoNameFull on N-Caltech, N-Cars, and N-ImageNet with our best settings.
    \label{table:best-zero-shot}
    }
    \vspace{-2.5mm}
\end{table}

\begin{table}[t]
    \centering
    \small
    \setlength{\tabcolsep}{4pt}
    \begin{tabular}{cccc|ccc}
        \toprule
        \textbf{Dataset} & \multicolumn{3}{c}{N-Caltech} & \multicolumn{3}{c}{N-ImageNet} \\
        $N$ ($\times 10^3$) & 15 & 20 & 25 & 50 & 70 & 80 \\
        \midrule
        0-shot Acc. & \textbf{69.87} & 69.67 & 69.33 & 20.04 & \textbf{20.78} & 20.61 \\
        10-shot Acc. & 84.98 & \textbf{85.62} & 84.93 & 27.83 & \textbf{28.63} & 27.62 \\
        \bottomrule
    \end{tabular}
    \vspace{\tablecapmargin}
    \caption{
    \textbf{Ablation on the event time window size $N$}. 
    We report zero-shot and ten-shot accuracy (\%) as measurements.
    \label{table:ablation-N}
    }
    \vspace{-2.5mm}
\end{table}

\begin{table}[t]
    \centering
    \small
    \setlength{\tabcolsep}{4pt}
    \begin{tabular}{cccc|ccc}
        \toprule
        \textbf{Dataset} & \multicolumn{3}{c}{N-Caltech} & \multicolumn{3}{c}{N-ImageNet} \\
        \textbf{Method} & Gray & R-B & Learn & Gray & R-B & Learn \\
        \midrule
        0-shot Acc. & \textbf{69.67} & 65.93 & - & \textbf{20.78} & 17.49 & - \\
        10-shot Acc. & 85.62 & 82.87 & 85.69 & 28.63 & 25.23 & 28.55 \\
        \bottomrule
    \end{tabular}
    \vspace{\tablecapmargin}
    \caption{
    \textbf{Ablation on the event histograms colorization methods}. 
    See text for implementation details of the ablated methods.
    \label{table:ablation-colorization}
    }
    \vspace{\tablemargin}
    \vspace{-1mm}
\end{table}

\subsection{Zero-Shot Classification}\label{sec:zero-shot-cls}


\heading{Results.}
\cref{table:best-zero-shot} presents the zero-shot classification accuracy of \algoNameFull. 
Without any in-domain training, our method achieves an accuracy of 69.67\% on N-Caltech which has 101 classes.
This proves the effectiveness of our event-to-frame conversion pipeline in bridging the RGB and event camera domains.
In addition, our model scores a higher 82.28\% accuracy on N-Cars which is captured in the real world, demonstrating its generalizability.
On the challenging N-ImageNet dataset, we achieve a lower accuracy of 20.78\% due to the lack of event domain knowledge.

\begin{figure*}[t]
    \vspace{\pagetopmargin}
    \vspace{-1mm}
    \centering
    \hspace*{-2mm}
    \begin{subfigure}{0.33\linewidth}
        \includegraphics[width=\linewidth]{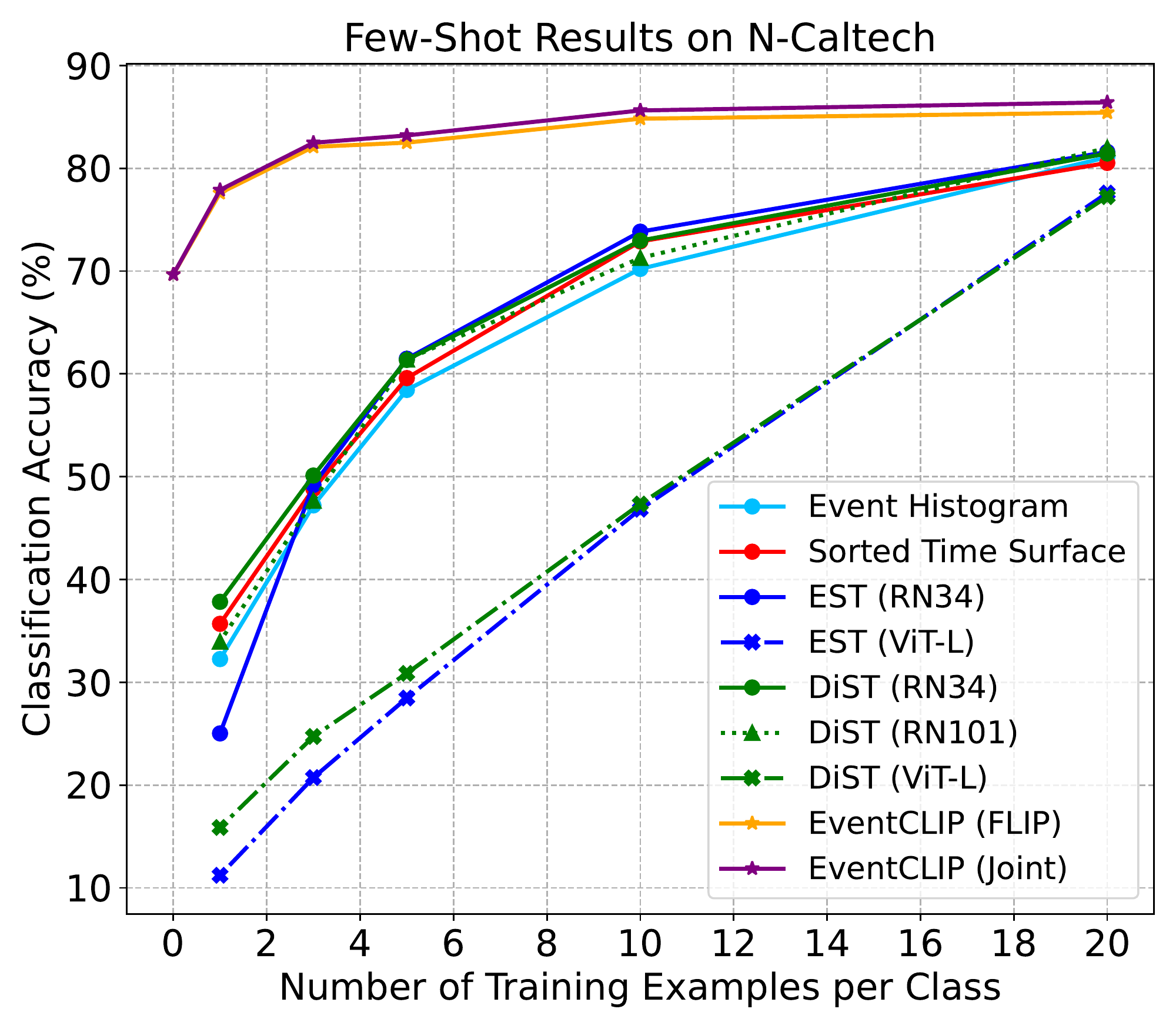}
    \end{subfigure}
    \hspace{-1mm}
    \begin{subfigure}{0.33\linewidth}
        \includegraphics[width=\linewidth]{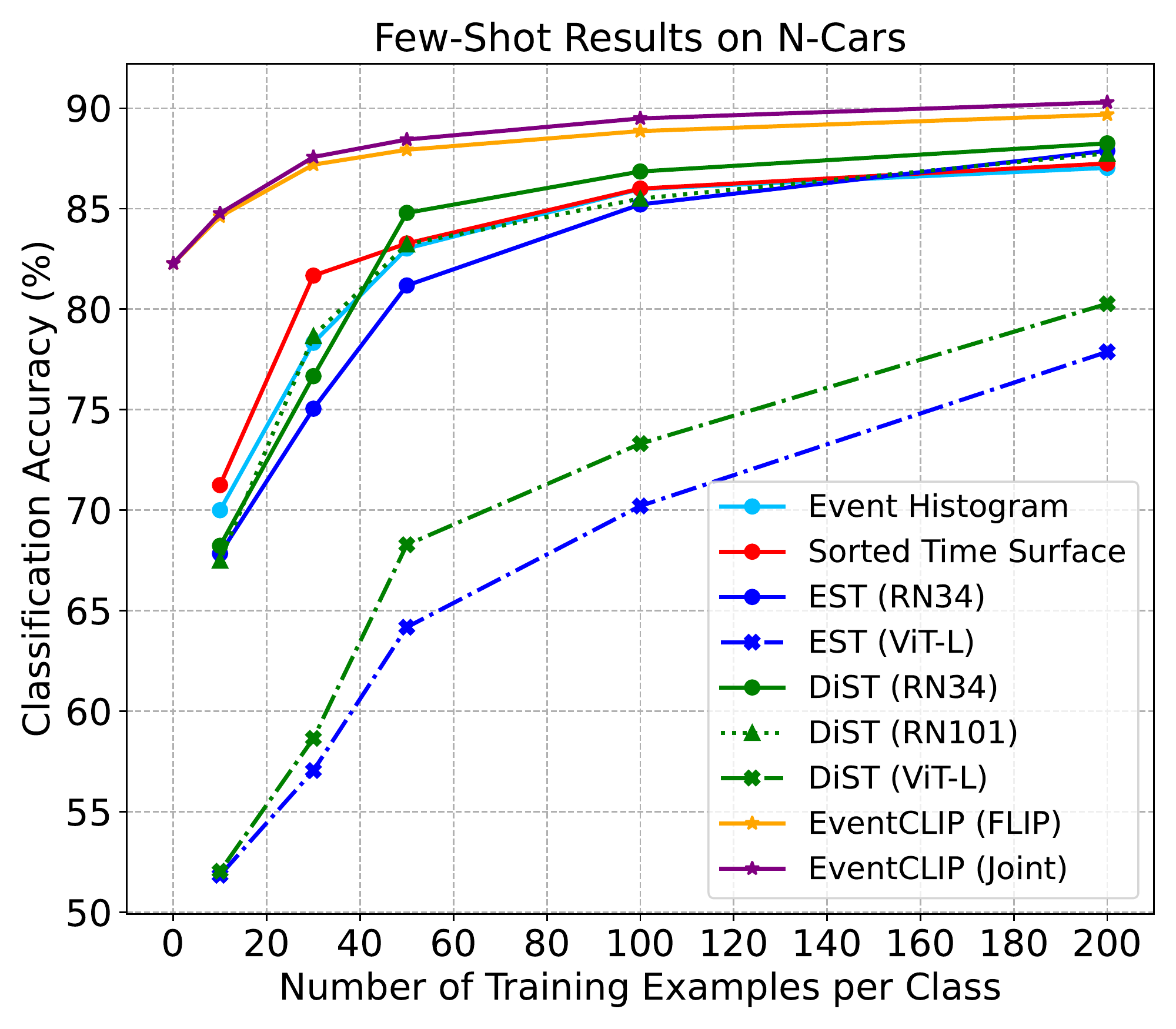}
    \end{subfigure}
    \hspace{-1mm}
    \begin{subfigure}{0.33\linewidth}
        \includegraphics[width=\linewidth]{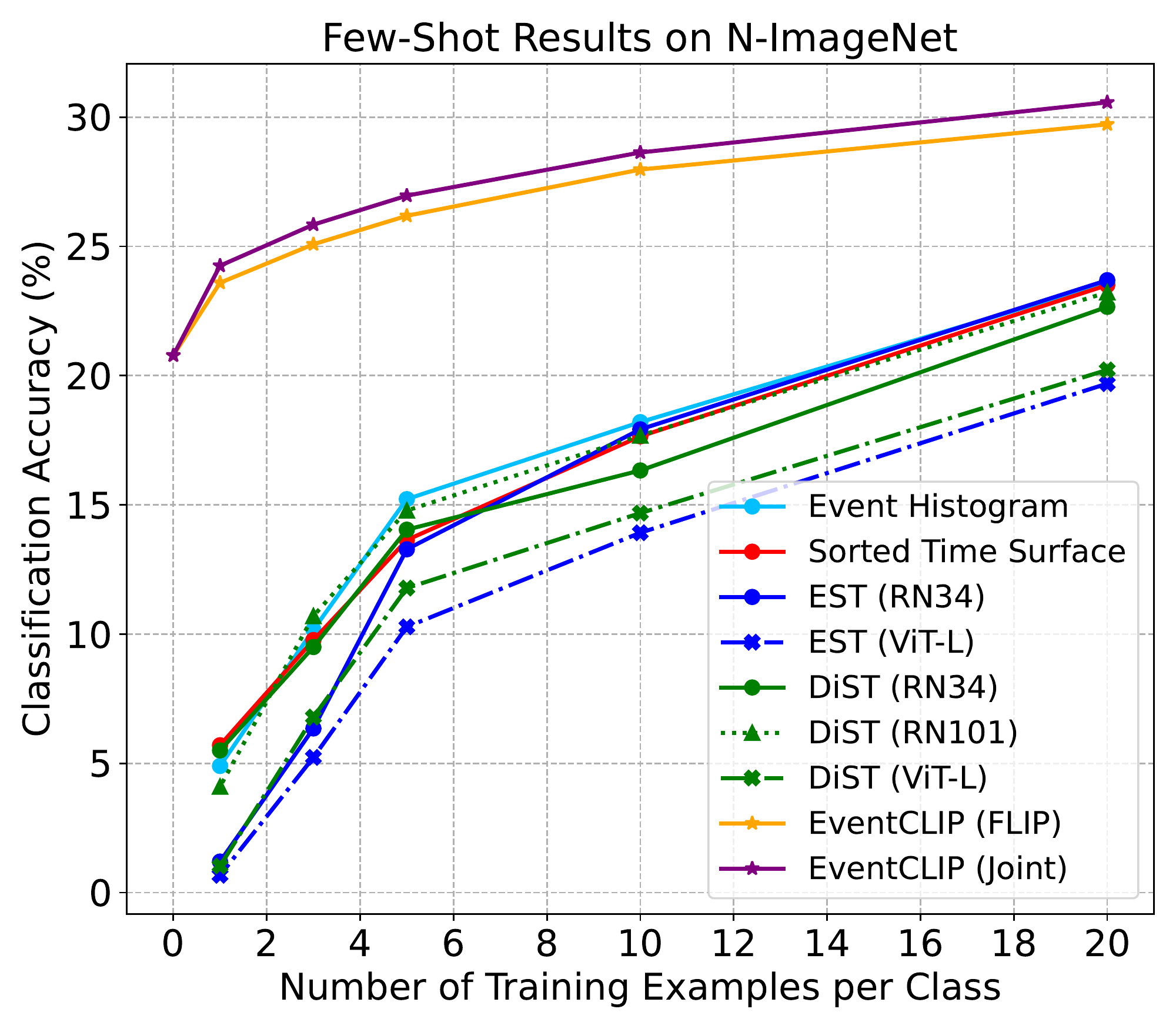}
    \end{subfigure}
    \vspace{\figcapmargin}
    \caption{
    \textbf{Few-shot classification accuracy} on N-Caltech, N-Cars, and N-ImageNet.
    All baselines build upon the ResNet34 backbone pre-trained on RGB ImageNet.
    For EST and DiST, we try additional variants with larger ResNet101 and ViT-L backbones.
    We also report \algoNameFull with another VLM FLIP~\cite{FLIP-kaiming}.
    Our method consistently outperforms state-of-the-art event-based classifiers across all settings.
    }
    \label{fig:few-shot-sota}
    \vspace{\figmargin}
\end{figure*}

\heading{Ablation Study.}
We first study the event time window size $N$ in \cref{table:ablation-N}.
Since event streams from N-Cars are generally sparse, we convert all events to one frame without ablation.
On N-Caltech, a smaller $N$ achieves the best zero-shot accuracy, but we select 20,000 since it strikes a better balance between zero- and few-shot results.
We need a higher $N = 70,000$ for optimal performance on N-ImageNet due to its higher camera resolution.
See Appendix~\ref{app:ablate-N} for more ablations on $N$.
\cref{table:ablation-colorization} ablates different ways of colorizing the event histogram.
We test the red-blue color map (dubbed R-B) commonly used to visualize events, which multiplies positive and negative event counts with [255, 0, 0] and [0, 0, 255], respectively.
It leads to much worse results because the color statistics of their converted images are distinct from the natural images CLIP is trained on.
We also design a learnable method by initializing the color map with two vectors and optimizing them jointly during training.
This leads to similar accuracies, but requires 10$\times$ the computation since it needs to backpropagate through the heavy ViT image encoder.
Overall, our gray-scale color map is efficient and reduces the domain gap of input to CLIP.

\begin{table}[t]
    \vspace{-0mm}
    \centering
    \small
    \setlength{\tabcolsep}{2.7pt}
    \begin{tabular}{ccccccc}
        \toprule
        \textbf{Model} & RN & RN$\times$4 & RN$\times$16 & RN$\times$64 & ViT-B & ViT-L \\
        \midrule
        Size (MB) & 244 & 403 & 631 & \textbf{1300} & 335 & 890 \\
        0-shot Acc. & 44.34 & 51.41 & 60.83 & 61.92 & 61.11 & \textbf{69.67} \\
        10-shot Acc. & 74.21 & 77.43 & 79.95 & 81.23 & 80.70 & \textbf{85.62} \\
        \bottomrule
    \end{tabular}
    \vspace{\tablecapmargin}
    \caption{
    \textbf{Ablation on the image encoder of CLIP on N-Caltech}.
    We test the ResNet50 (RN) family and Vision Transformers (ViT-B/16, ViT-L/14).
    Model sizes and accuracies are reported.
    \label{table:ablation-visual-encoder}
    }
    \vspace{-2.5mm}
\end{table}

\begin{table}[t]
    \centering
    \small
    \setlength{\tabcolsep}{2pt}
    \begin{tabular}{lcc}
        \toprule
        \textbf{Prompt} & 0-shot Acc. & 10-shot Acc. \\
        \midrule
        A photo of a [CLASS] & 66.57 & 82.18 \\
        An event camera photo of [CLASS] & 64.73 & 78.34 \\
        An edge map of a [CLASS] & 68.70 & 84.07 \\
        A sketch image of a [CLASS] & 69.64 & 85.16 \\
        A point cloud image of a [CLASS] & 69.67 & \textbf{85.62} \\
        \text{[Learnable Tokens]} + [CLASS] & - & 85.37 \\
        \bottomrule
    \end{tabular}
    \vspace{\tablecapmargin}
    \caption{
    \textbf{Ablation on the text prompt templates on N-Caltech}.
    ``Learnable Tokens" trains 16 context vectors as in CoOp~\cite{CoOp}.
    \label{table:ablation-prompt}
    }
    \vspace{\tablemargin}
    \vspace{-1mm}
\end{table}

One important property of foundation models is the scalability of their performance with model size.
We study this effect using different image encoders of CLIP in \cref{table:ablation-visual-encoder}.
Within the ResNet~\citep{ResNet} and the ViT~\citep{ViT} family, \algoNameFull achieves higher accuracy as the model size grows.
Notably, ViT-L significantly outperforms RN50$\times$64, despite having much fewer parameters.
The reason might be the converted event frames mostly capture the object boundary, and are thus biased towards the shape information.
Studies have shown that CNNs are usually texture-biased~\citep{CNNTextureBias}, while ViTs are better at processing shape information~\citep{ViTShapeBias1,ViTShapeBias2}.
This observation may serve as future guidelines in designing event-based vision architectures.
\cref{table:ablation-prompt} compares the template used to construct text prompts.
The template ``\texttt{A photo of a [CLASS]}" commonly used in 2D vision tasks achieves 66.57\% zero-shot accuracy.
Simply prefixing ``photo" with ``event camera" leads to worse results as CLIP is not trained on event data.
Instead, we explicitly describe the visual property of event frames.
Since events are mostly triggered by object boundaries, ``edge map" and ``sketch image" both lead to better results.
Surprisingly, describing event frames with ``point cloud image" achieves the highest accuracy, which aligns with previous works that treat raw events as spatio-temporal points~\citep{SpaceTimeEventCloud}.
We also tried prompt tuning with learnable textual tokens~\citep{CoOp}, which achieves similar performance.
However, it trains 5$\times$ slower as it requires backpropagation through the heavy text encoder.


\subsection{Few-Shot Classification}\label{sec:few-shot-cls}

\heading{Settings.}
We experiment with 1, 3, 5, 10, 20 shots on N-Caltech and N-ImageNet.
Since N-Cars only has two categories, we multiply the number of shots by 10.
We test \algoNameFull with four variants of adapters:
\textbf{(i)} PointCLIP's MLP visual adapter~\citep{PointCLIP},
\textbf{(ii)} our proposed Transformer visual adapter,
\textbf{(iii)} the Classifier Tuning text adapter~\citep{WiSE-FT},
and \textbf{(iv)} a joint adapter combining (ii) and (iii).
We also test our generality by replacing CLIP with another VLM FLIP~\cite{FLIP-kaiming}.

\heading{Results.}
We first compare \algoNameFull using joint adapter with baselines in \cref{fig:few-shot-sota}.
All baselines with ResNet34 backbone achieve similar performance across datasets, which aligns with previous observation~\citep{DiST-N-IN}.
In contrast, \algoNameFull achieves significantly higher few-shot accuracy. 
Our 20-shot accuracy on N-Caltech (85.62\%) surpasses EST trained on the entire N-Caltech (81.7\%) by around 4\%.
Notice that, all baselines are initialized with backbones pre-trained on RGB ImageNet, which is the source for creating N-ImageNet.
Still, \algoNameFull achieves consistently higher accuracy across all numbers of shots.
Overall, the results prove that CLIP's large-scale pre-training learns generalizable representations, enabling quick adaptation to the new event camera domain with limited training data.

For a fair comparison, we also evaluated baselines with larger backbones ResNet101 and ViT-L.
Naively fine-tuning ViTs on limited data usually leads to severe overfitting~\citep{ViT}, so we adopt the state-of-the-art data-efficient ViT training strategy~\citep{DEIT}.
As shown in the figures, DiST with ResNet101 leads to similar performance, while ViT-L results in much worse accuracy even with the advanced training strategy.
This indicates the superiority of our \algoNameFull framework as it scales well with larger models.

Finally, we replace CLIP with another VLM FLIP~\cite{FLIP-kaiming} (dubbed \textit{\algoNameFull (FLIP)}).
This variant achieves slightly lower performance compared to using CLIP, as FLIP also achieves lower zero-shot accuracy than CLIP on RGB image datasets.
Nonetheless, \algoNameFull with FLIP still outperforms baselines, showing its generality with base VLMs.

\begin{figure}[t]
    \vspace{\pagetopmargin}
    \vspace{-2mm}
    \centering
    \includegraphics[width=0.99\linewidth]{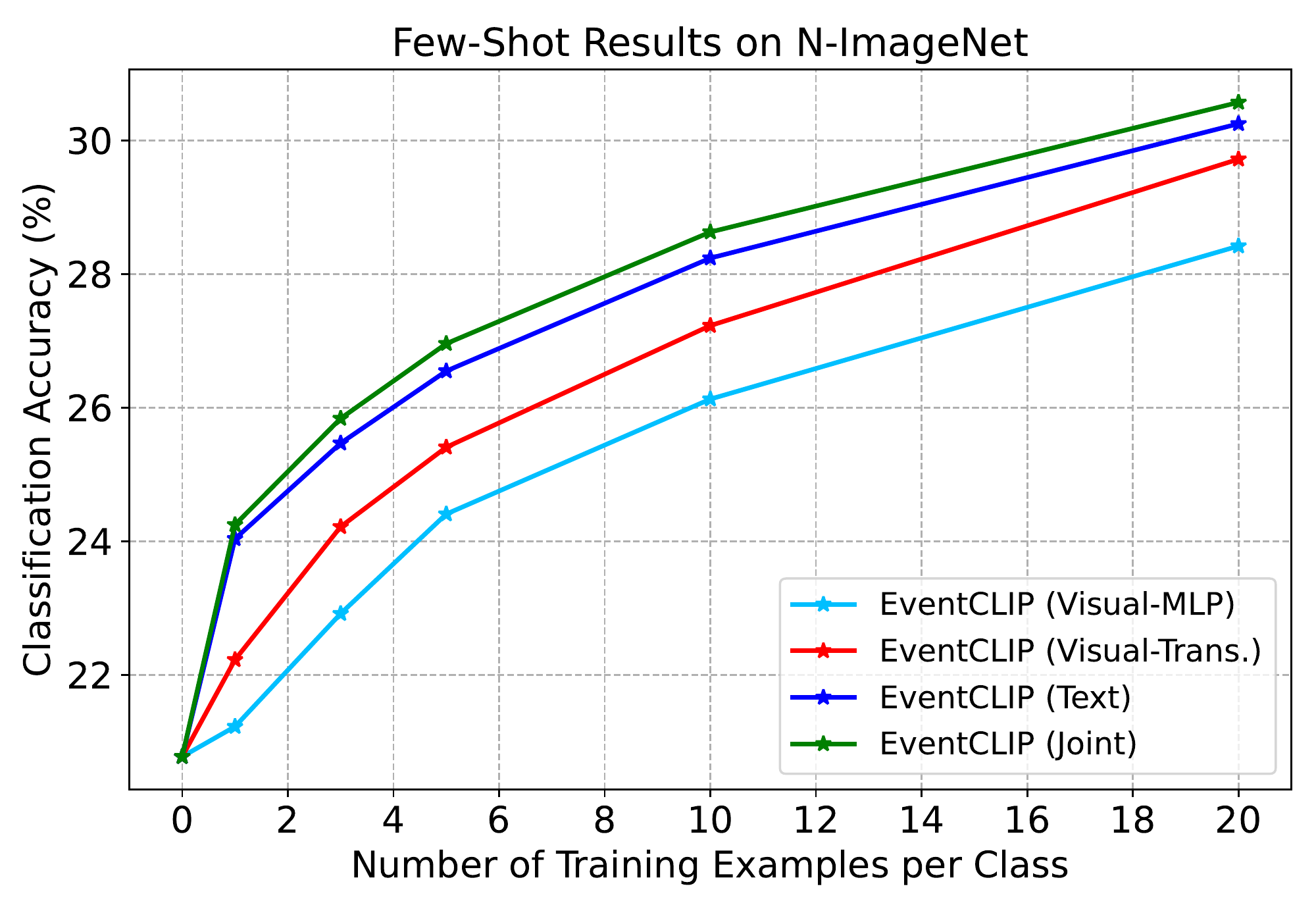}
    \vspace{\figcapmargin}
    \vspace{-1mm}
    \caption{
    \textbf{Ablation on feature adapters} of \algoNameFull on N-ImageNet.
    Trans. stands for the Transformer-based visual adapter.
    }
    \label{fig:few-shot-ablation}
    \vspace{\figmargin}
    \vspace{-1mm}
\end{figure}

\heading{Ablation Study.}
\cref{fig:few-shot-ablation} compares the performance of \algoNameFull with four feature adapters on N-ImageNet.
The Transformer-based visual adapter consistently outperforms the MLP-based counterpart by a sizeable margin, showing the importance of its permutation-equivariant property.
Interestingly, adapting the text features along yields clearly better results than adapting the image features along, which is distinct from prior works on images~\cite{CLIP-Adapter,Tip-Adpater}.
This may be due to the larger domain gaps event frames pose compared to images.
Finally, adapting both branches leads to the best performance without suffering from overfitting.
See Appendix~\ref{app:more-adapter-ablation} for more ablations on N-Caltech and N-Cars.

\subsection{Fine-Tuning \algoNameFull}\label{sec:fine-tune}

\heading{Settings.}
Few-shot feature adaptation is able to improve the model accuracy with minor computation and time overhead.
However, it still underperforms event-based classifiers trained on the entire dataset.
To show that \algoNameFull can also achieve state-of-the-art performance when more data are available, we propose to fine-tune the CLIP's image encoder jointly with the text feature adapter.

\heading{Baseline.}
We compare with concurrent work E-CLIP~\citep{E-CLIP}, which also uses CLIP but requires \textit{paired RGB images} with events during training.
E-CLIP is the current state-of-the-art method using full datasets.
For a fair comparison, we adopt CLIP with the same ViT-B/16 backbone as they do.

\heading{Results.}
\cref{table:compare-e-clip} presents the fine-tuning results of \algoNameFull.
Following the protocol of E-CLIP, we also report the accuracy in the few-shot setting.
On N-Caltech, we achieve better results in the low data regime.
With more data, \algoNameFull is competitive with E-CLIP.
We hypothesize that this is because N-Caltech is extremely class-imbalanced where some categories only have 20 samples.
As a result, the paired RGB images provide much useful information for E-CLIP.
This is verified by the results on N-ImageNet with abundant samples per class.
Without using additional data, \algoNameFull still outperforms E-CLIP consistently over all settings, achieving new state-of-the-art performance. 

\begin{table}[t]
    \vspace{\pagetopmargin}
    \vspace{-0mm}
    \small
    \begin{subtable}{1.0\linewidth}
        \centering
        \setlength{\tabcolsep}{3.5pt}
        \begin{tabular}{ccccccc}
            \toprule
            Data per Class & 1 & 2 & 5 & 10 & 20 & All \\
            \midrule
            E-CLIP & 66.72 & 75.87 & 82.35 & 86.92 & 90.51 & 93.89 \\
            \textbf{\algoNameFull} & \textbf{75.82} & \textbf{78.86} & \textbf{83.57} & 87.42 & 90.41 & 93.57 \\
            \bottomrule
        \end{tabular}
        \caption{Fine-tuning results on N-Caltech}
        \label{table:e-clip-ncaltech}
    \end{subtable}
    \\
    \begin{subtable}{1.0\linewidth}
        \vspace{1mm}
        \centering
        \setlength{\tabcolsep}{3.5pt}
        \begin{tabular}{ccccccc}
            \toprule
            Data per Class & 1 & 2 & 5 & 10 & 20 & All \\
            \midrule
            E-CLIP & 22.22 & 26.85 & 28.70 & 30.56 & 35.11 & 51.85 \\
            \textbf{\algoNameFull} & \textbf{24.39} & \textbf{27.23} & \textbf{31.12} & \textbf{34.24} & \textbf{38.28} & \textbf{53.20} \\
            \bottomrule
        \end{tabular}
        \caption{Fine-tuning results on N-ImageNet}
        \label{table:e-clip-nin}
    \end{subtable}
    \vspace{\tablecapmargin}
    \vspace{-3mm}
    \caption{
    \textbf{Fine-tuning results}.
    We compare \algoNameFull with E-CLIP~\citep{E-CLIP} which fine-tunes the CLIP image encoder.
    We report results under the few-shot setting or trained on all data in the dataset.
    \label{table:compare-e-clip}
    }
    \vspace{\tablemargin}
    \vspace{-5mm}
\end{table}

\begin{table*}[t]
    \vspace{\pagetopmargin}
    \vspace{-1mm}
    \centering
    \small
    \setlength{\tabcolsep}{5pt}
    \begin{tabular}{l|c|ccccc|cccc|c}
        \toprule
        \textbf{Data Variation} & None & \multicolumn{5}{c}{Trajectory} & \multicolumn{4}{c}{Brightness} & Average \\
        \midrule
        \textbf{Variant ID.} & Orig. & 1 & 2 & 3 & 4 & 5 & 6 & 7 & 8 & 9 & All \\
        \midrule
        \textbf{Event Histogram} & 47.73 & 43.73 & 33.72 & 37.69 & 24.56 & 35.24 & 20.89 & 29.68 & 36.33 & 34.56 & 32.93 \\
        + DiST & 51.67 & 48.02 & 38.18 & 43.16 & 27.56 & 40.02 & 25.19 & 34.22 & 40.63 & 38.83 & 37.31 \\
        + Ev-TTA (online) & - & 44.94 & 44.63 & 43.31 & 41.48 & 43.46 & 26.89 & 34.71 & 43.86 & 43.42 & 40.86 \\
        + Ev-TTA (offline) & - & 48.64 & \underline{48.01} & \textbf{47.24} & \underline{44.49} & \textbf{47.06} & \underline{30.08} & \underline{38.34} & 47.37 & 46.58 & \underline{44.20} \\
        \midrule
        + \algoNameFull (0-shot) & 50.03 & 48.49 & 43.33 & 41.57 & 37.90 & 40.14 & 25.72 & 34.28 & 44.33 & 44.65 & 40.05 \\
        + \algoNameFull (20-shot) & 51.68 & \underline{51.06} & 46.58 & 43.63 & 42.59 & 42.94 & 27.64 & 37.18 & \underline{47.65} & \underline{46.93} & 42.91 \\
        + \algoNameFull (100-shot) & \textbf{52.72} & \textbf{51.98} & \textbf{48.82} & \underline{44.98} & \textbf{44.75} & \underline{44.79} & \textbf{30.26} & \textbf{38.84} & \textbf{49.68} & \textbf{48.20} & \textbf{44.70} \\
        \specialrule{1pt}{2pt}{2pt}
        \textbf{Sorted Time Surface} & 47.90 & 44.33 & 33.50 & 40.17 & 23.72 & 37.19 & 21.57 & 30.31 & 36.63 & 35.18 & 33.62 \\
        + DiST & 51.56 & 47.92 & 37.92 & 43.84 & 27.07 & 40.64 & 25.38 & 34.35 & 40.49 & 38.87 & 37.39 \\
        + Ev-TTA (online) & - & 46.02 & 45.29 & 45.91 & 42.53 & 43.90 & 26.70 & 36.17 & 45.00 & 45.22 & 41.86 \\
        + Ev-TTA (offline) & - & 49.58 & \underline{47.67} & \textbf{48.36} & \textbf{45.59} & \textbf{46.72} & \underline{30.07} & \underline{39.30} & \underline{48.24} & \underline{47.76} & \underline{44.81} \\
        \midrule
        + \algoNameFull (0-shot) & 50.25 & 48.78 & 43.12 & 44.03 & 37.24 & 41.66 & 26.23 & 35.21 & 44.39 & 43.82 & 40.50 \\
        + \algoNameFull (20-shot) & \underline{52.23} & \underline{51.27} & 46.65 & 46.07 & 42.38 & 44.22 & 28.22 & 37.95 & 48.04 & 47.22 & 43.56 \\
        + \algoNameFull (100-shot) & \textbf{52.85} & \textbf{52.78} & \textbf{48.92} & \underline{47.44} & \underline{44.68} & \underline{45.55} & \textbf{30.53} & \textbf{39.68} & \textbf{49.63} & \textbf{49.19} & \textbf{45.38} \\
        \specialrule{1pt}{2pt}{2pt}
        \textbf{DiST} & 48.43 & 45.17 & 36.58 & 42.28 & 26.57 & 38.70 & 24.39 & 32.76 & 38.99 & 37.37 & 35.89 \\
        + Ev-TTA (online) & - & 46.32 & 46.05 & 46.57 & 43.23 & 44.58 & 28.05 & 36.98 & 46.03 & 45.64 & 42.61 \\
        + Ev-TTA (offline) & - & 48.53 & \underline{47.75} & \textbf{48.38} & \textbf{45.35} & \textbf{47.26} & 31.02 & 39.07 & 48.19 & \underline{47.66} & \underline{44.80} \\
        \midrule
        + \algoNameFull (0-shot) & 50.53 & 49.36 & 44.47 & 45.12 & 37.83 & 43.15 & 28.01 & 36.79 & 45.72 & 44.58 & 41.67 \\
        + \algoNameFull (20-shot) & \underline{52.28} & \underline{51.42} & 47.53 & 46.77 & 42.34 & 45.05 & 29.62 & 39.05 & \underline{49.00} & 47.52 & 44.26 \\
        + \algoNameFull (100-shot) & \textbf{53.12} & \textbf{52.45} & \textbf{49.01} & \underline{47.78} & \underline{43.85} & \underline{46.22} & 31.00 & \textbf{39.92} & \textbf{49.77} & \textbf{48.74} & \textbf{45.42} \\
        \bottomrule
    \end{tabular}
    \vspace{\tablecapmargin}
    \caption{
    \textbf{Classification accuracy on N-ImageNet and its robustness variants}.
    The numbers of baseline event-based classifiers and Ev-TTA are copied from \citet{Ev-TTA}.
    Average is computed over 9 variants.
    Best results are bold and the second-best results are underlined.
    \label{table:robust-ensemble}
    }
    \vspace{\tablemargin}
    \vspace{-5mm}
\end{table*}

\subsection{Robust Event Classification}\label{sec:robust-cls}

\heading{Settings.}
Our goal is to evaluate whether the large-scale 2D pre-trained knowledge in CLIP is complementary to existing event-based classifiers.
Therefore, we perform model ensemble using 0-, 20-, and 100-shot \algoNameFull (few-shot trained with the joint feature adapter) with baselines trained on \textit{the entire N-ImageNet} by simply averaging the predicted class logits.
We test the original and the ensembled model on the normal test set of N-ImageNet, as well as its 9 variants~\citep{DiST-N-IN} which are captured under out-of-distribution camera motions and lighting conditions.

\heading{Baselines.}
We adopt pre-trained weights of baselines from the official codebase of N-ImageNet.
EST is excluded since there is no pre-trained weight available.
We report the accuracy of model ensemble with DiST, as DiST is currently the most robust event-based classifier.
We also compare with Ev-TTA~\citep{Ev-TTA}, which is designed specifically for event domain adaptation.
Importantly, Ev-TTA requires access to \textit{data in the new domain} to perform test-time adaptation to update model weights.
It has an online version where novel testing data are used only once and an offline version where the out-of-distribution data are used multiple times.

\heading{Results.}
As shown in \cref{table:robust-ensemble}, model ensemble with 0-shot \algoNameFull already increases the accuracy of baselines by more than 5\%, which is higher than ensemble with a fully trained DiST.
This indicates that pre-trained CLIP contains information that cannot be effectively learned from event camera datasets only.
Such information complements pre-trained event-based classifiers, making them more robust against data corruption.
Compared to ensemble with DiST trained on the entire N-ImageNet, ensemble with 20-shot \algoNameFull improves the average accuracy on robustness variants by more than 5\%.
It is worth noting that DiST achieves 48.43\% accuracy on the original test set, while our 20-shot \algoNameFull scores a much lower 30.57\%.
Besides, ensemble DiST with a worse performing \algoNameFull still greatly improves the model performance on both the original and the robustness variants of N-ImageNet.

Next, we compare our ensemble method with the sophisticated test-time adaptation method Ev-TTA, which requires access to additional out-of-distribution data.
Surprisingly, \algoNameFull trained on 20 samples per category (less than 2\% of all training data) outperforms the online version of Ev-TTA on most of the subsets. 
With 8\% of training data, our 100-shot \algoNameFull outperforms the offline version of Ev-TTA, achieving new state-of-the-art robustness results.
See Appendix~\ref{app:qual-results} for additional qualitative analysis.


\subsection{Label-Free Event Recognition}\label{sec:unsup-learning}

\heading{Settings and Baseline.}
We follow Ev-LaFOR~\citep{Ev-LaFOR} to perform \textit{fully unsupervised learning} on a 100-class subset of N-ImageNet.
Please refer to their paper for details on the N-ImageNet (Mini) split.
Ev-LaFOR reconstructs images from events and uses CLIP for classification.
It has two variants, and we compare with the better-performing one that uses unlabeled events and \textit{RGB images}.
For a fair comparison, we adopt CLIP with the same ViT-B/32 backbone.

\heading{Results.}
\algoNameFull with ViT-B/32 backbone achieves a zero-shot accuracy of 27.08\% on the test set of N-ImageNet (Mini).
Naive self-training degrades it to 26.43\%.
With our consistency-based labeling process, we can improve the performance to 35.26\%, surpassing Ev-LaFOR's accuracy of 31.28\% by around 4\%.
See Appendix~\ref{app:semi-sup-eventclip} for additional results under the semi-supervised learning setting.

\section{Conclusion}

In this paper, we propose \algoNameFull which adapts CLIP for open-set event recognition.
With event-to-frame conversion, we successfully transfer CLIP's 2D pre-trained knowledge to the event camera domain.
To further enhance the performance, we develop lightweight adapters to refine the pre-trained CLIP embeddings.
Moreover, \algoNameFull can be employed to improve the robustness of existing classifiers via model ensemble, or learn from unlabeled data with self-training.
Our work opens up new possibilities to apply recent advances in foundation models to event-based vision.
We discuss the limitations of this work in Appendix~\ref{app:limitations}.

\subsection*{Acknowledgments}

We acknowledge the support of the Natural Sciences and Engineering Research Council of Canada (NSERC).
This research was enabled in part by support provided by Compute Ontario (\url{www.computeontario.ca}), the Digital Research Alliance of Canada (\url{https://alliancecan.ca/}), the Province of Ontario, and the Government of Canada through CIFAR, and companies sponsoring the Vector Institute (\url{www.vectorinstitute.ai/partnerships/current-partners/}).
We would also like to acknowledge Vector Institute for computation support.

\noindent We would like to thank Yang Zheng for the help with data preparation, Xiaoshi Wu for discussions about CLIP models, and Mathias Gehrig, Yash Kant, Xuanchi Ren, Liquan Wang for valuable discussions and support.





{
    \small
    \bibliographystyle{ieeenat_fullname}
    \bibliography{references}
}

\newpage
\appendix
\section{Additional Related Work}\label{app:more-related-work}

\heading{Deep Learning for Event-based Classification.}
Depending on the utilization of the sparsity and asynchronous nature of event data, existing event-based classifiers can be mainly categorized into two classes~\citep{EventVisionSurvey}, namely, synchronous and asynchronous methods.
Synchronous models aggregate events to a grid-based representation, and then use standard modules such as Convolutional Neural Networks (CNNs) to process it~\citep{STS,EH,HATS-N-Cars,EST,Video2Event,EventGraftNet,DiST-N-IN}.
Significant efforts have been made to achieve efficient and expressive event-to-frame conversion~\citep{BinEvImg,EH,STS}.
Recently, EST~\citep{EST} has achieved state-of-the-art results in classification with an end-to-end learnable event-to-frame conversion pipeline.
As a remedy for robustness in the presence of noise, DiST~\citep{DiST-N-IN} proposes to suppress noisy events leveraging their spatio-temporal relationships, which is proved effective under camera motion and lighting variations.

On the other hand, asynchronous networks~\citep{SNN-N-MNIST,PhasedLSTM,TruthNet,HOTS,EventSparseConv,MatrixLSTM,SlideGCN,AEGNN} have been developed to address the computational latency inherent in grid-based methods, which directly apply Spiking Neural Networks or Graph Neural Networks to raw event inputs.
However, these methods still consistently underperform synchronous methods across datasets~\citep{N-Caltech,HATS-N-Cars,DiST-N-IN}.
As our primary goal is to achieve high accuracy instead of efficiency, we adopt the former category of methods as our baselines in the experiments.

\heading{Bridging Frame-based and Event-based Vision.}
Inspired by the success of classical computer vision, several works have introduced techniques from frame-based vision to process event data.
Some papers focus on reconstructing natural images from events, and then apply conventional deep models on the converted frames~\citep{E2VID,E2VID-PAMI,FireNet,BetterE2VID,E2VID-Transformer,Ev-LaFOR}.
However, they introduce computational overhead which is at odds with event cameras' low-latency nature.
Another line of work tries to simulate event data from existing RGB image datasets, where ground-truth annotations can be automatically obtained from labeled frames~\citep{DAVIS-Sim,ESIM,V2E,Video2Event,EventGAN}.
The drawback is the large Sim2Real gap of the synthesized events such as unrealistic camera motions and the absence of sensor noises.
The most relevant works to ours are methods that transfer knowledge learned from RGB images to event-based models~\citep{EventGraftNet,Evdistill,ESS_EvSegFromImg,WormholeLearningRSS,WormholeLearningICRA,EventDA}.
However, they either require paired recordings of image and event data, or massive labels in the image domain.
In this work, we utilize CLIP pre-trained on RGB image-text pairs for data-efficient event-based classification.
Our method converts events into frames via simple counting, and directly applies CLIP for zero-shot classification.
We can further boost its performance via few-shot fine-tuning, without the need for paired RGB images or large amounts of labels.

\heading{CLIP-based Few-Shot Transfer Learning.}
Transfer learning aims to leverage models trained on large-scale datasets to help learning on data-scarce tasks.
In the field of event-based object recognition, existing grid-based methods have also utilized models pre-trained on RGB images from ImageNet as their backbones to improve performance via fine-tuning~\citep{EST,DiST-N-IN,EH,STS}.
Trained on large-scale image-text pairs, CLIP~\citep{CLIP} has shown great potential in learning transferable representations for various downstream tasks.
To further enhance the few-shot accuracy of CLIP, one line of work~\citep{CoOp,CoCoOp,KgCoOp} proposes to insert learnable text tokens to perform task-specific prompt tuning.
CLIP-Adapter~\citep{CLIP-Adapter}, Tip-Adapter~\citep{Tip-Adpater}, and WiSE-FT~\citep{WiSE-FT} instead learn lightweight adapters over CLIP features.

In addition to 2D image classification, CLIP has also been extended to 2D detection~\citep{ViLD,Detic}, segmentation~\citep{DenseCLIP1,DenseCLIP2}, and video analysis~\citep{ActionCLIP,VideoCLIPPrompting}.
Our work is inspired by PointCLIP~\citep{PointCLIP,PointCLIP_V2}, which projects point clouds to multi-view images and performs zero-shot and few-shot shape recognition with CLIP.
Different from PointCLIP, events only capture the boundary information of objects compared to the complete surfaces presented in point clouds.
Also, we design a Transformer-based adapter for event temporal information fusion, while PointCLIP simply uses an MLP since their multi-view projections follow a fixed order.

\section{Details on N-ImageNet Robustness Variants}\label{app:NIN-variants}

Here, we provide more information about the robustness variants of N-ImageNet \citep{DiST-N-IN} test sets.
The original training and testing set (it is actually the validation set, but we call it test set for simplicity) of N-ImageNet are both captured with a Samsung DVS Gen3 \citep{DVE_Gen3} event camera moving around the screen under the same environmental conditions.
To test event-based classifiers' robustness against variations in the data capture process, the authors create 9 variants of the same test set.
Variants 1-5 change the camera motions used to trigger events, where different moving directions, frequencies, and amplitudes of the camera trajectory are employed.
Variants 6-9 alternate the lighting conditions of the environment, such as extremely low or high illuminations.
Overall, these variations cause a large degradation in the performance of existing event-based classifiers.
See Appendix~\ref{app:qual-results} for visualizations of some data variants.

\section{Implementation Details}\label{app:implement-details}

In our few-shot experiments, all models including baselines and \algoNameFull are trained on the same subset of data.
All reported results are averaged over three runs, and we empirically find that the performance variation is small.

\heading{Baselines.}
\textit{EST}~\citep{EST} is the state-of-the-art method on N-ImageNet which utilizes learnable kernels to convert raw events into grid-based representations.
\textit{Event Histogram}~\citep{EH} converts the event counts into a two-channel image grouped by their polarity.
\textit{Sorted Time Surface}~\citep{STS} adopts the sorted indices of event timestamps to ensure durability against camera speed changes.
\textit{DiST}~\citep{DiST-N-IN} is specifically designed to improve the robustness against event camera noise and corruptions, which filters noise with local statistics.
It achieves state-of-the-art results on the robustness benchmark of N-ImageNet variants.

We adopt the online official implementation of EST\footnote{\href{https://github.com/uzh-rpg/rpg\_event\_representation\_learning}{https://github.com/uzh-rpg/rpg\_event\_representation\_learning}}~\citep{EST} and DiST\footnote{\href{https://github.com/82magnolia/n\_imagenet}{https://github.com/82magnolia/n\_imagenet}}~\citep{DiST-N-IN}.
The implementation of Event Histogram and Sorted Time Surface are also adopted from the DiST codebase.
We re-train all the models with their default settings on each dataset, but decrease the learning rate and number of training epochs when observing severe overfitting.
ResNet34~\citep{ResNet} pre-trained on the RGB version of ImageNet~\citep{ImageNet} is adopted as their backbones, and fine-tuned jointly under the few-shot setting.
As shown in the experiments, we tried DiST with pre-trained ResNet50 and ResNet101 backbones, but did not observe clear improvement in accuracy.
We also tried pre-trained ViT~\citep{ViT} models.
Even with state-of-the-art data-efficient training strategy~\citep{DEIT}, the joint fine-tuning suffers from severe overfitting, leading to results even worse than ResNet backbones.
We hypothesize that this is because vision transformers are data-hungry, and thus are not suitable for the few-shot learning setting.
To evaluate the model ensemble performance on the robustness variants of N-ImageNet, we directly use the pre-trained weights from the official release.

\heading{Data Augmentation.}
Following DiST~\citep{DiST-N-IN}, we use random jittering, random horizontal flip along the spatial dimension, and random reverse along the temporal dimension as data augmentations.
We tried other event augmentations such as random event dropping, and random cropping over both spatial and temporal dimensions, but did not observe clear improvement.
Since we convert events to 3-channel RGB frames, \algoNameFull additionally benefits from the well-studied RGB image augmentation literature.
We apply RandAugment~\citep{RandAugment} \footnote{We use the implementation from \href{https://pytorch.org/vision/stable/generated/torchvision.transforms.RandAugment.html}{torchvision}} to the converted event frames during training.
For each loaded event stream, we apply the same set of operations to all frames converted from this data.
RandAugment consistently improves the performance of \algoNameFull on N-Caltech and N-ImageNet, while bringing little gain on N-Cars.
Finally, the resulting frames are resized and center cropped to 224 $\times$ 224 following CLIP.

\heading{Few-shot \algoNameFull.}
We adopt the pre-trained weights of CLIP from their official online release\footnote{\href{https://github.com/openai/CLIP}{https://github.com/openai/CLIP}}.
For \algoNameFull with the Transformer-based visual adapter, we stack 2 standard Transformer encoder modules~\citep{Attention}, with a token size equal to $256$ and 4 heads.
We choose the pre-LN Transformer variant~\citep{Pre-LN-Trans} as it leads to more stable training and less overfitting.
To ensure the permutation-equivariant property, we do not apply positional encoding to the Transformer input following \citet{XuanchiRetriever} and \citet{SlotFormer}.
For the MLP-based visual adapter baseline, we adopt the best-performing setting from PointCLIP~\citep{PointCLIP}, which concatenates image features $F = \{\bm{f}_i\}_{i=1}^M$ to extract a global feature, and fuses with per-frame features via residual connections.
For the text adapter, we treat the text features $W=\{\bm{w}_i\}_{i=1}^K$ as the weight of the fully-connected layer in a classifier, and update it via gradient descent.
When applying the visual adapter only, we set the residual ratio $\alpha$ to $0.5$, while we use $\alpha = 0.8$ to further alleviate overfitting when training two adapters jointly.
On N-ImageNet, we always use $\alpha = 0.95$ as we observe severe overfitting.

We train all models with the Adam~\citep{Adam} optimizer for 100 epochs on N-Caltech and N-ImageNet, and 50 epochs on N-Cars.
We use a batch size of 32 on N-Caltech and N-Cars, and 128 on N-ImageNet.
When the number of training data is smaller than 32, \eg N-Cars under the 10-shot learning setting only has 20 samples, we set the batch size as the number of data available.
On N-Caltech and N-Cars, when applying the visual and text adapter separately, we set the peak learning rate as $2 \times 10^{-4}$ and $1 \times 10^{-3}$ for them, respectively.
When training them jointly, we set the peak learning rate as $2 \times 10^{-4}$.
Besides, we divide the learning rate by a factor of 10 when training on N-ImageNet to present overfitting.
We also adopt a linear learning rate warmup schedule during the first 5$\%$ of training steps, and decay the learning rate to $0$ in a cosine schedule.
We do not use any weight decay or gradient clipping as we did not find them useful in preliminary experiments.

\heading{Fine-tuning \algoNameFull.}
We fine-tune the image encoder of CLIP, while keeping the text encoder frozen.
We use the same hyper-parameters and training settings as few-shot \algoNameFull, except that we use a 10$\times$ smaller learning rate on CLIP's image encoder.
When fine-tuned on the entire dataset, we train shorter for 50 epochs.
Besides tuning all the model parameters, we also tried other parameter-efficient fine-tuning methods such as only tuning the bias terms, only tuning LayerNorm, and LoRA~\citep{LoRA}, but observed worse performance compared to naive fine-tuning.

\heading{Robust event classification with \algoNameFull.}
We average the class logits of \algoNameFull with class logits predicted by baseline event-based classifiers.
We search for the weight factors to balance the two terms on the normal validation set, and fix it for all the remaining robustness variants.

\heading{Learning from unlabeled data with \algoNameFull.}
There are two settings in this task: fully unsupervised learning where no labels are available, and semi-supervised learning where we have a few labeled samples per class.
In the unsupervised setting, we use the zero-shot \algoNameFull to generate pseudo labels.
In the semi-supervised setting, we first train \algoNameFull on the labeled data similar to few-shot learning, and then use it to generate pseudo labels.
We choose horizontal flip and temporal flip as TTA methods.
Given an unlabeled event $\mathcal{E}$, we first generate four versions of it $\{\mathcal{E}_i\}_{i=1}^4$ by applying the augmentations combinatorially.
Then, we run \algoNameFull to predict class probabilities $\{\bm{p}_i \in \mathbb{R}^K\}_{i=1}^4$ ($K$ is the number of classes), and get the class labels $\{c_i = \argmax_{c} \bm{p}_i\}_{i=1}^4$.
We discard examples with an inconsistent $\{c_i\}_{i=1}^4$.
To further enhance the label quality, we select examples with confidence scores higher than $\tau$, \ie $\max(\bm{p}_i) > \tau, \forall\ i \in \{1, 2, 3, 4\}$.
Finally, we take the top-$k$ most confident examples from each category to form the pseudo-labeled training set.
After obtaining the training data, we follow few-shot \algoNameFull to train our model on it with the joint feature adapter.

For hyper-parameters, we choose $k = 30$ in both settings.
For $\tau$, we need to use a very high value of $0.999$ in unsupervised learning, as the zero-shot \algoNameFull is over-confident in its predictions.
After few-shot adaptation, \algoNameFull is better calibrated, and we can use a lower value of $0.5$.
The model performance is insensitive to it here.

\section{Additional Experimental Results}\label{app:more-exp-results}

\begin{figure}[t!]
    \vspace{\pagetopmargin}
    \centering
    \begin{subfigure}{0.95\linewidth}
        \includegraphics[width=\linewidth]{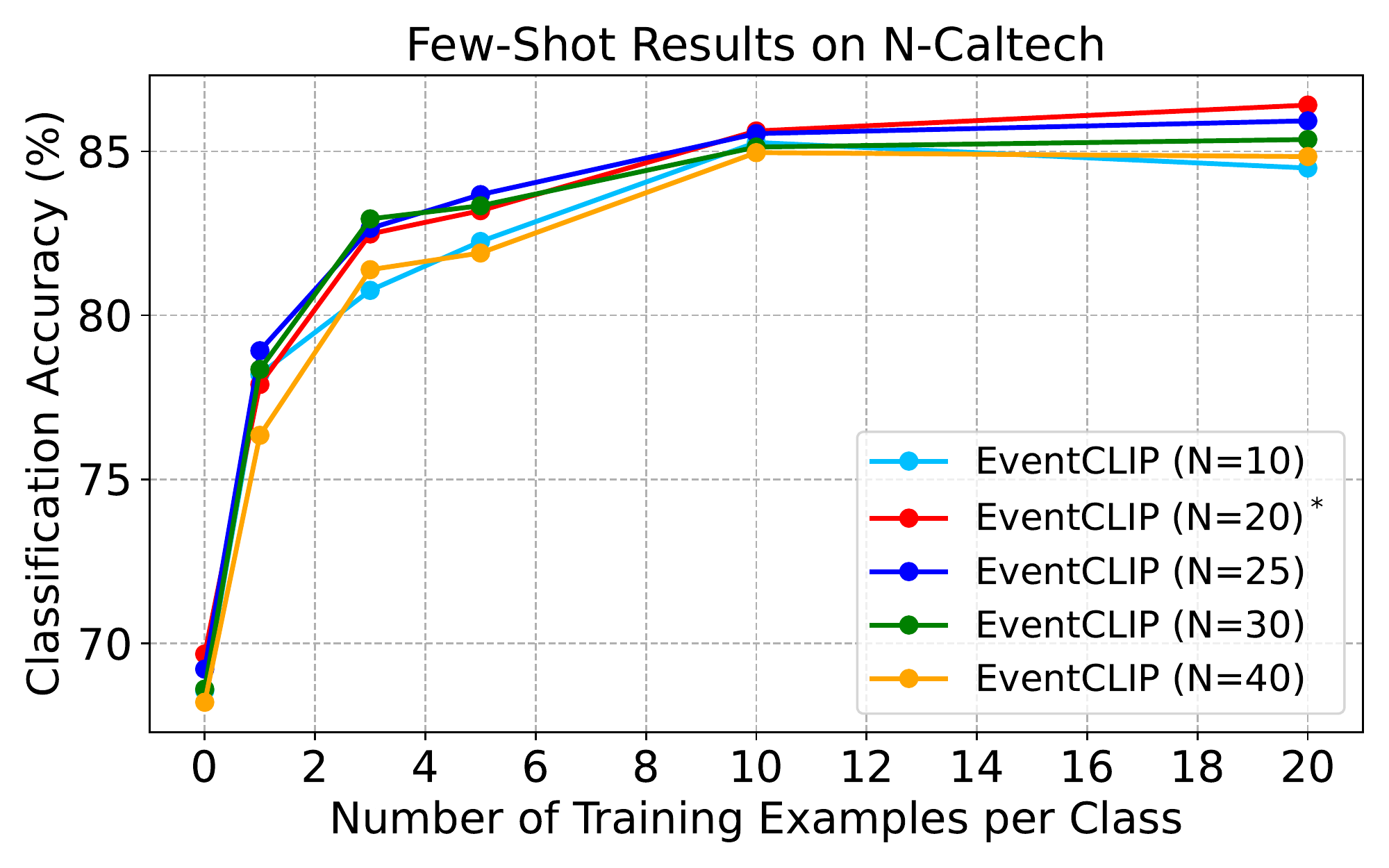}
    \end{subfigure}
    \\
    \begin{subfigure}{0.95\linewidth}
        \includegraphics[width=\linewidth]{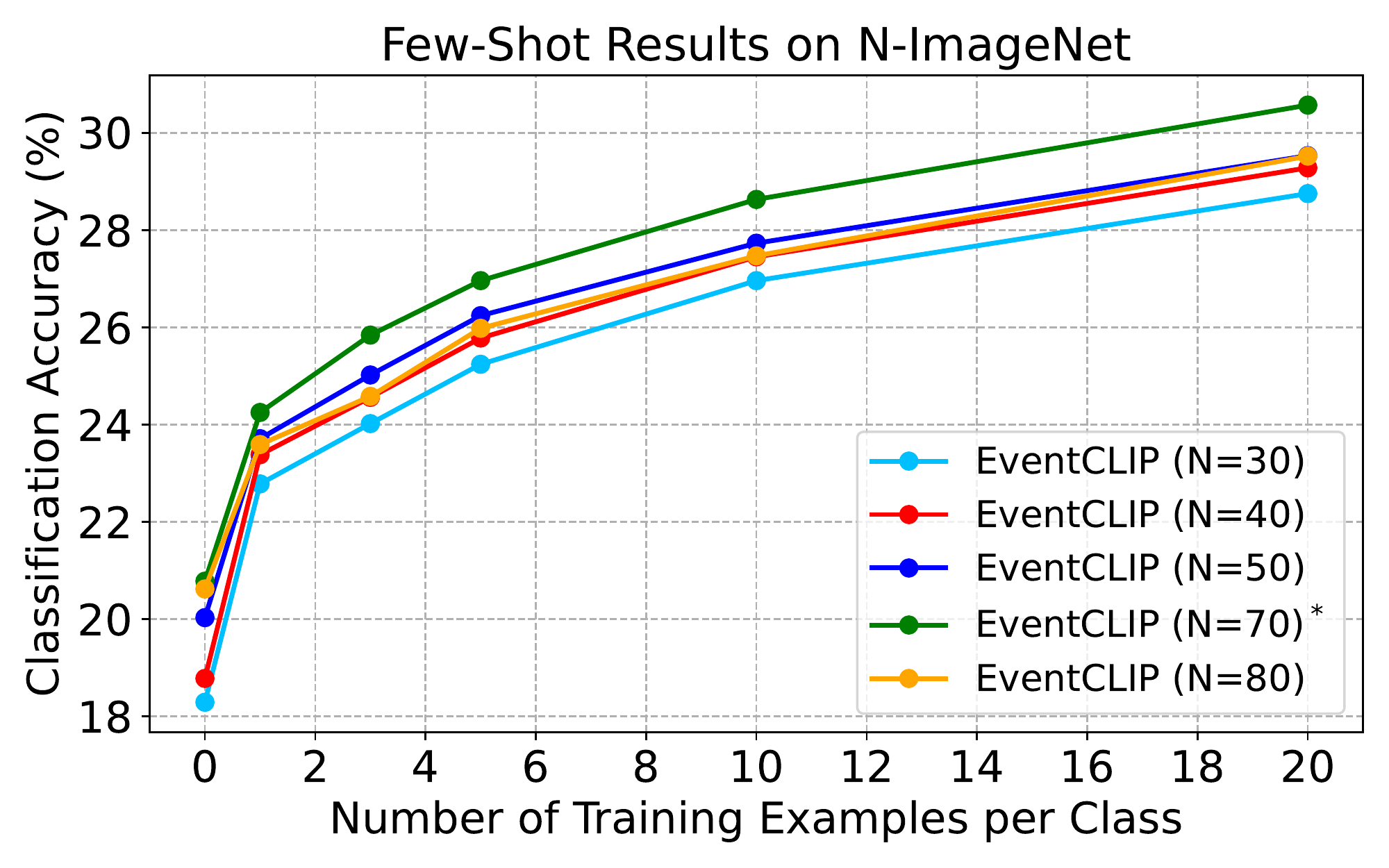}
    \end{subfigure}
    \vspace{-1mm}
    \vspace{\figcapmargin}
    \caption{
    \textbf{Ablation on event time window size} $N$ ($\times 10^3$) on N-Caltech and N-ImageNet.
    Entry with $^*$ is the optimal setting.
    }
    \label{app-fig:ablate-N}
    \vspace{\figmargin}
    \vspace{0mm}
\end{figure}

\subsection{Ablation Study on Event Time Window Size $N$}\label{app:ablate-N}

Robust event-to-frame conversion is an active research field in event-based vision.
In this paper, we adopt the simple event histogram representation as it already gives good performance.
To resist camera and object motion changes, we convert every $N$ event into one frame.
The optimal $N$ varies across datasets, as they are often captured by event cameras with different resolutions.
We study the effect of $N$ on few-shot classification accuracy in \cref{app-fig:ablate-N}.
Overall, \algoNameFull is not sensitive to $N$ within a reasonable range, as the accuracy difference is smaller than 2$\%$ in most cases.
As a future direction, one can explore better event representations to further improve \algoNameFull's performance.

\begin{figure}[t]
    \vspace{\pagetopmargin}
    \vspace{-0mm}
    \centering
    \begin{subfigure}{0.95\linewidth}
        \includegraphics[width=\linewidth]{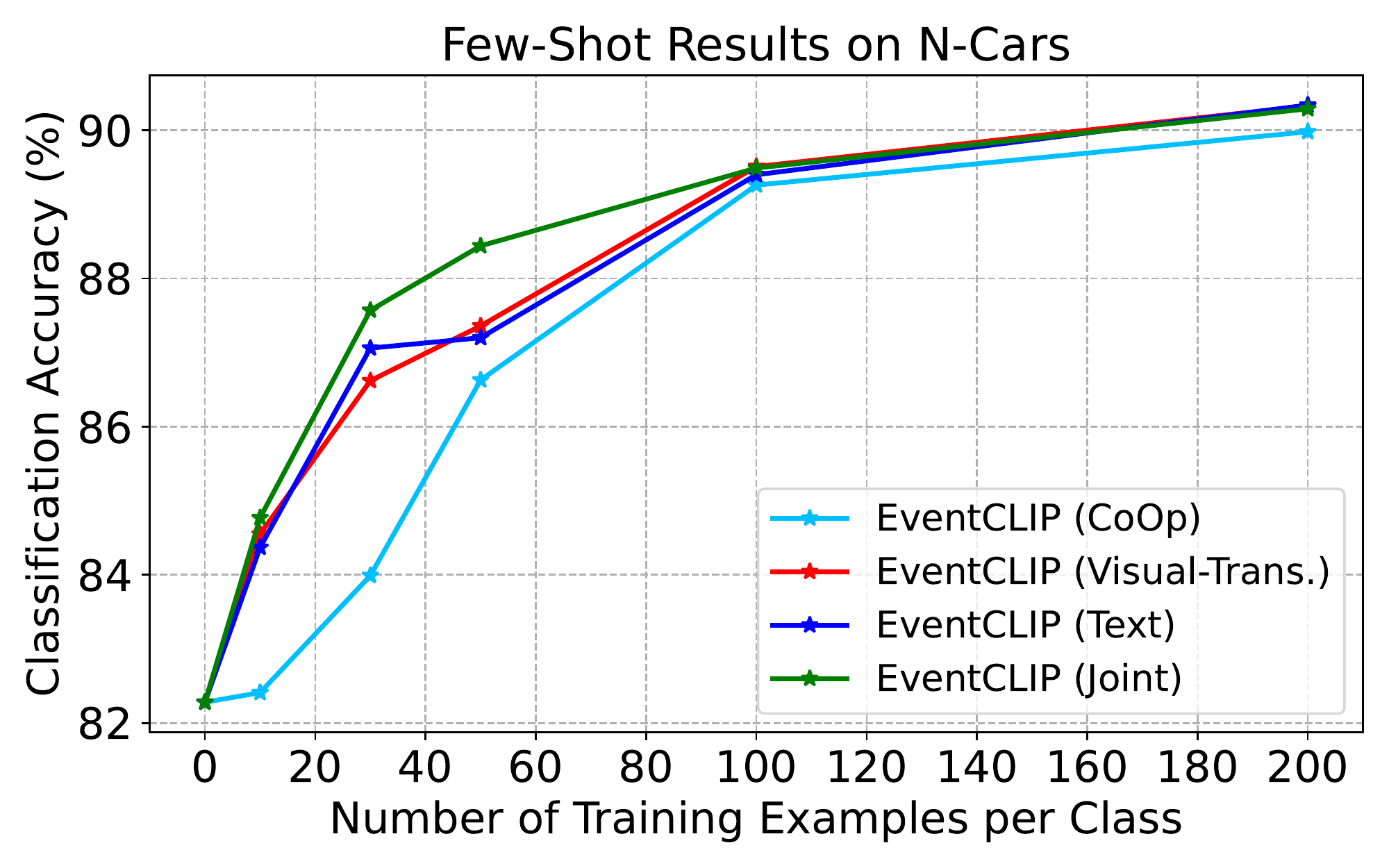}
    \end{subfigure}
    \\
    \begin{subfigure}{0.95\linewidth}
        \includegraphics[width=\linewidth]{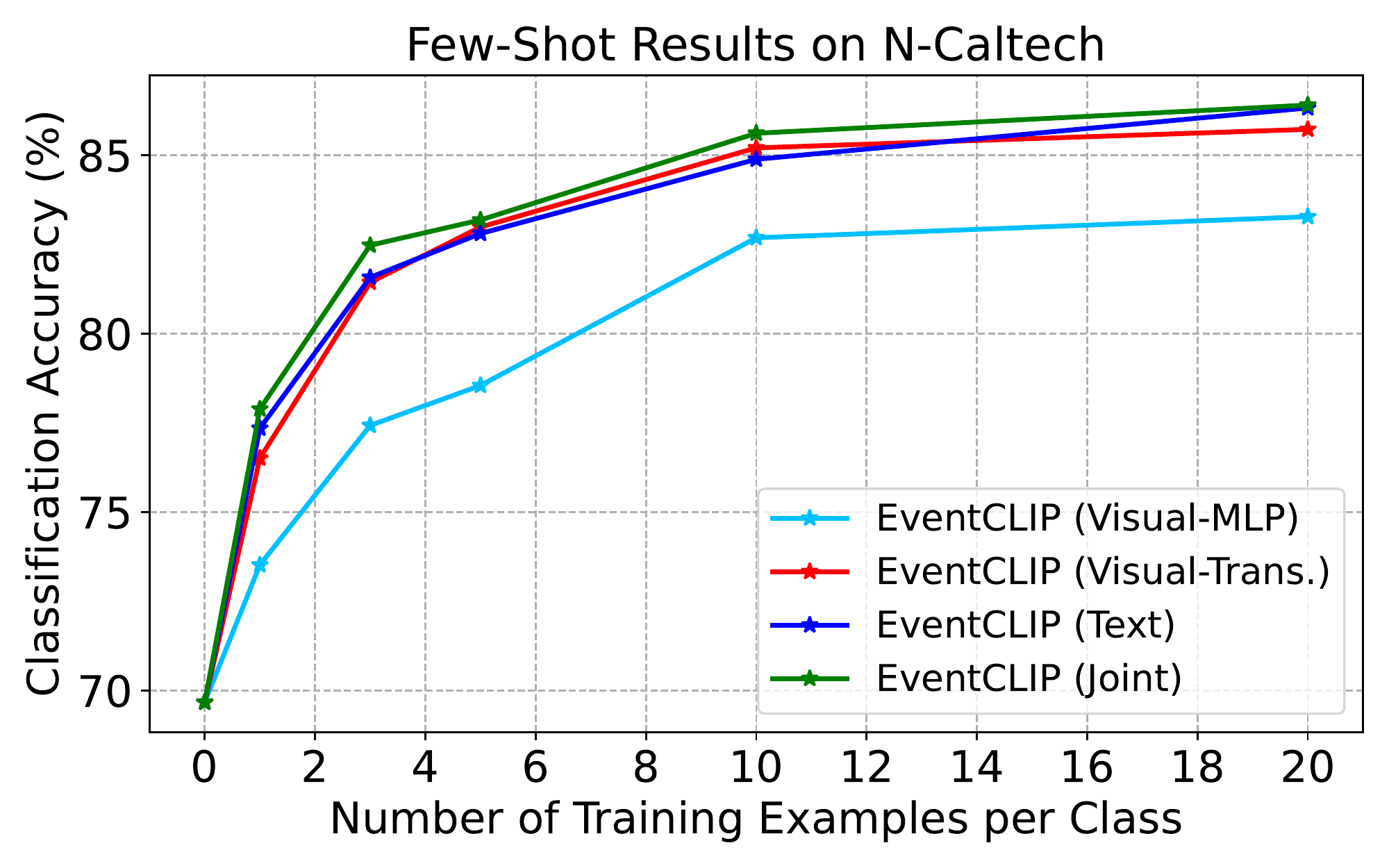}
    \end{subfigure}
    \vspace{\figcapmargin}
    \vspace{-0mm}
    \caption{
    \textbf{Ablation on feature adapters} on N-Cars and N-Caltech.
    Trans. stands for the Transformer-based visual adapter.
    CoOp means the text adapter based on prompt tuning.
    }
    \label{app-fig:ablate-adapter}
    \vspace{\figmargin}
\end{figure}

\subsection{Ablation Study on Feature Adapters}\label{app:more-adapter-ablation}

We perform ablation studies on the feature adapters on N-Cars and N-Caltech.
On N-Cars, we test an additional text adapter with prompt tuning (dubbed as CoOp~\citep{CoOp}), which learns context vectors in text prompts instead of using pre-defined templates.
\cref{app-fig:ablate-adapter} presents the few-shot accuracy of different adapters.
CoOp adapter is consistently worse than other adapters, while consuming much more resources as it requires backpropagation through CLIP's text encoder.
Similar to N-Caltech, the joint feature adapter performs better in the low-shot scenarios, and the gap becomes negligible when more data are provided.
Overall, we recommend users choose the text feature adapter as a starting point, since it achieves competitive performance, while requiring less computation and hyper-parameter tuning.
Instead, the joint adapter can achieve the best performance after tuning.

\subsection{Qualitative Results on Robust Classification}\label{app:qual-results}

\begin{figure*}[t]
    \vspace{-8mm}
    \begin{minipage}{.5\linewidth}
        \centering
        \begin{subfigure}{0.47\linewidth}
            \includegraphics[width=\linewidth]{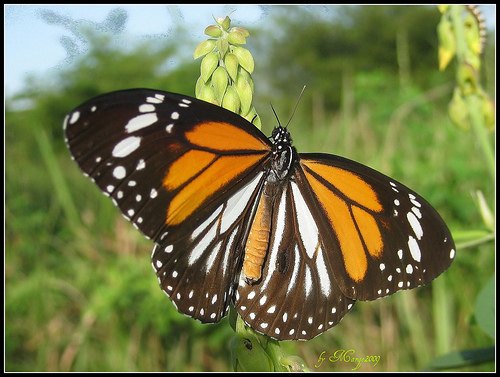}
            \caption{RGB image}
        \end{subfigure}\hspace{-2mm}
        \begin{subfigure}{0.47\linewidth}
            \includegraphics[width=\linewidth]{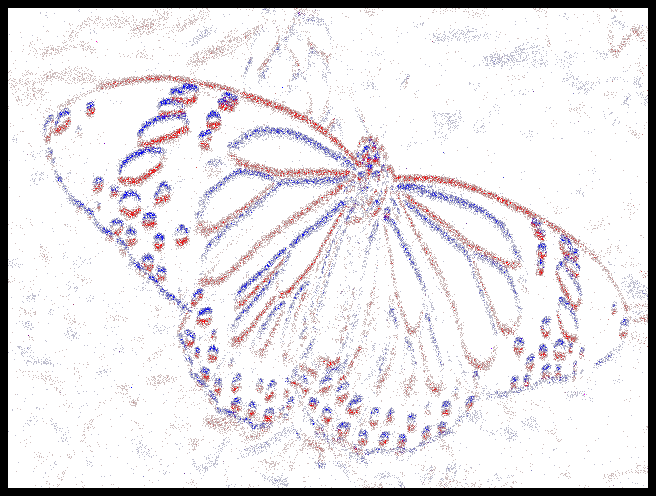}
            \caption{Original}
        \end{subfigure}
        \\
        \begin{subfigure}{0.315\linewidth}
            \includegraphics[width=\linewidth]{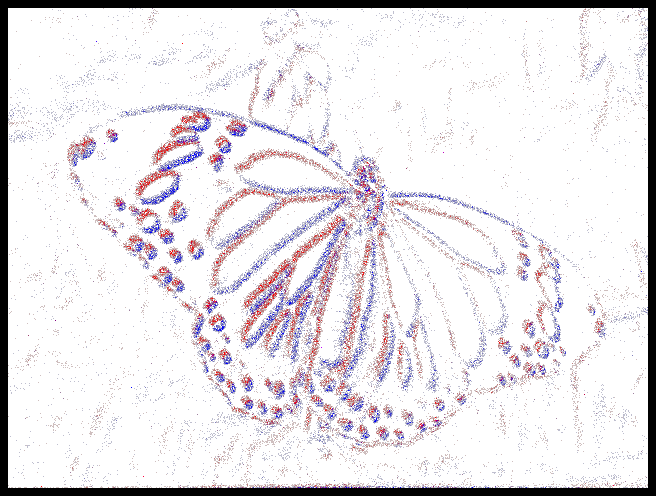}
            \caption{Variant 2}
        \end{subfigure}\hspace{-2mm}
        \begin{subfigure}{0.315\linewidth}
            \includegraphics[width=\linewidth]{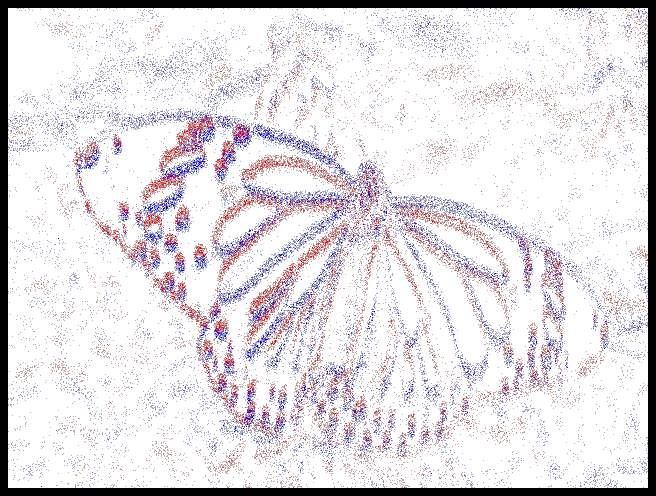}
            \caption{Variant 2}
        \end{subfigure}\hspace{-2mm}
        \begin{subfigure}{0.315\linewidth}
            \includegraphics[width=\linewidth]{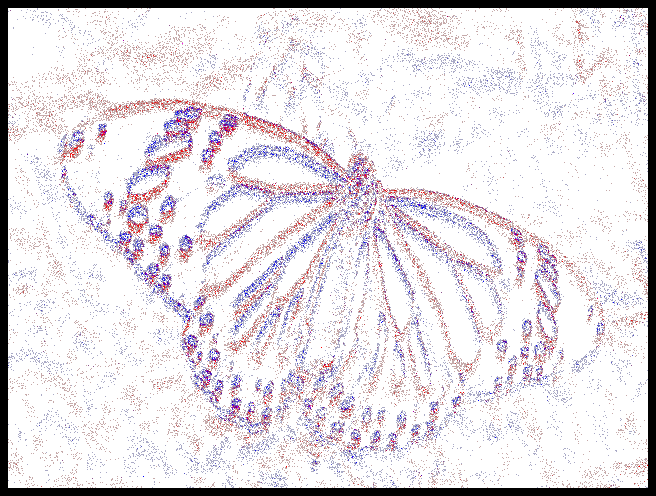}
            \caption{Variant 5}
        \end{subfigure}
        \\
        \begin{subfigure}{0.315\linewidth}
            \includegraphics[width=\linewidth]{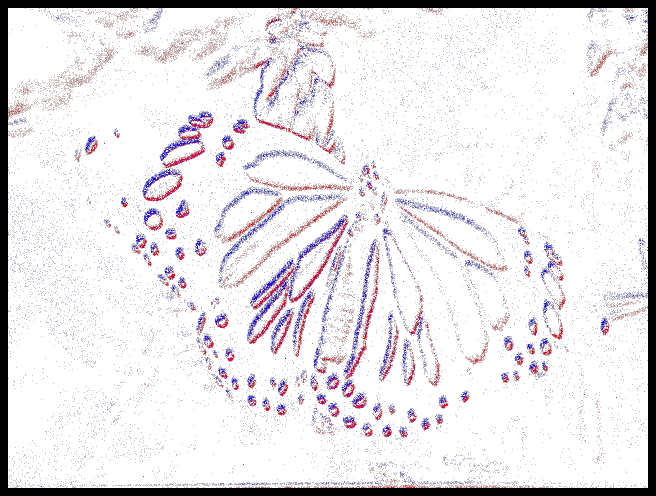}
            \caption{Variant 6}
        \end{subfigure}\hspace{-2mm}
        \begin{subfigure}{0.315\linewidth}
            \includegraphics[width=\linewidth]{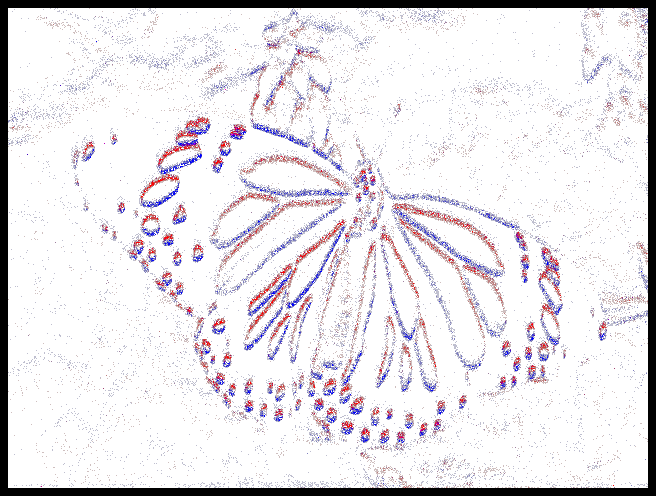}
            \caption{Variant 7}
        \end{subfigure}\hspace{-2mm}
        \begin{subfigure}{0.315\linewidth}
            \includegraphics[width=\linewidth]{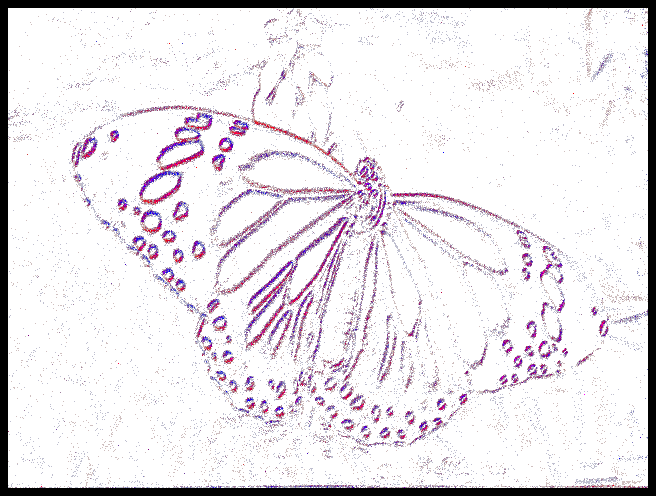}
            \caption{Variant 9}
        \end{subfigure}
    \end{minipage}
    \begin{minipage}{.5\linewidth}
        \centering
        \setcounter{subfigure}{0} 
        \begin{subfigure}{0.47\linewidth}
            \includegraphics[width=\linewidth]{imgs/qual/monarch-butterfly_16162/ILSVRC2012_val_00012137.JPEG}
            \caption{RGB image}
        \end{subfigure}\hspace{-2mm}
        \begin{subfigure}{0.47\linewidth}
            \includegraphics[width=\linewidth]{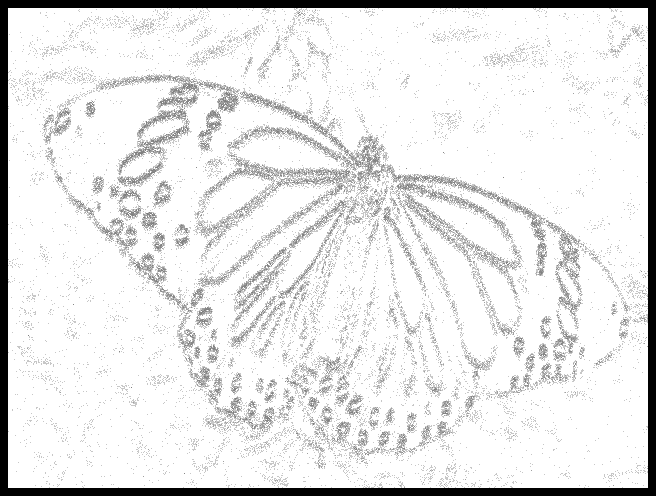}
            \caption{Original}
        \end{subfigure}
        \\
        \begin{subfigure}{0.315\linewidth}
            \includegraphics[width=\linewidth]{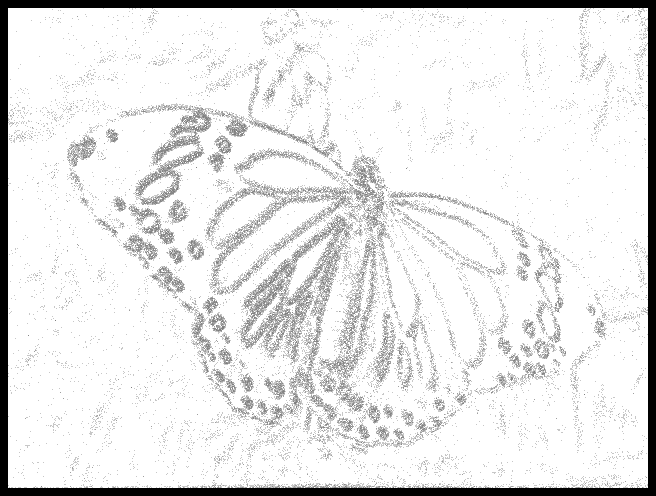}
            \caption{Variant 2}
        \end{subfigure}\hspace{-2mm}
        \begin{subfigure}{0.315\linewidth}
            \includegraphics[width=\linewidth]{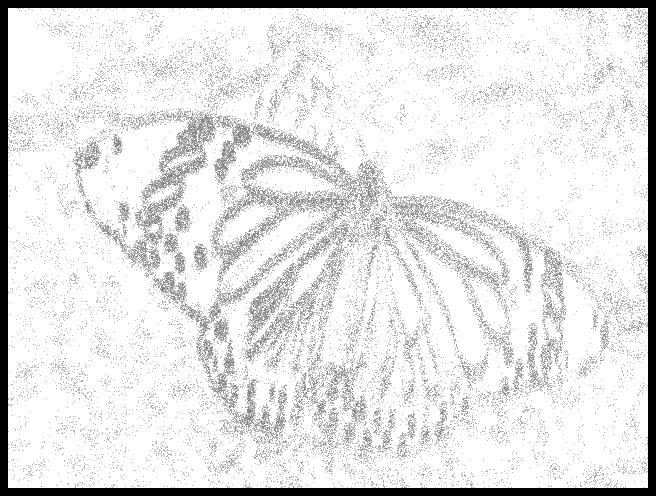}
            \caption{Variant 2}
        \end{subfigure}\hspace{-2mm}
        \begin{subfigure}{0.315\linewidth}
            \includegraphics[width=\linewidth]{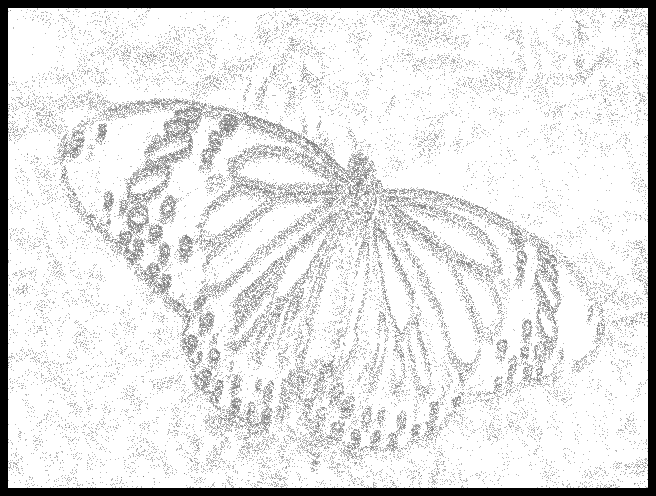}
            \caption{Variant 5}
        \end{subfigure}
        \\
        \begin{subfigure}{0.315\linewidth}
            \includegraphics[width=\linewidth]{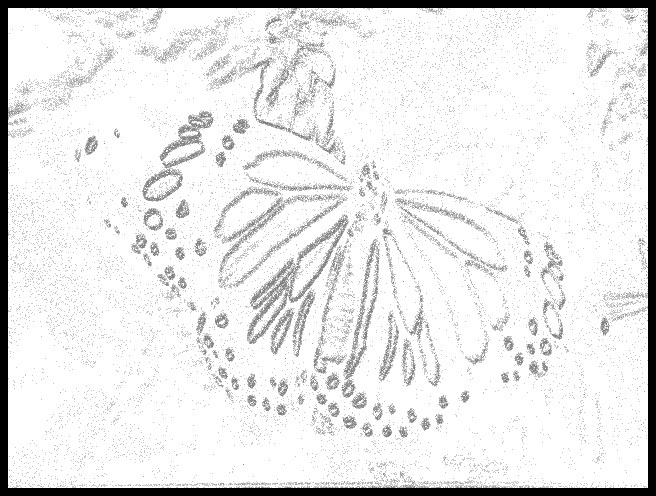}
            \caption{Variant 6}
        \end{subfigure}\hspace{-2mm}
        \begin{subfigure}{0.315\linewidth}
            \includegraphics[width=\linewidth]{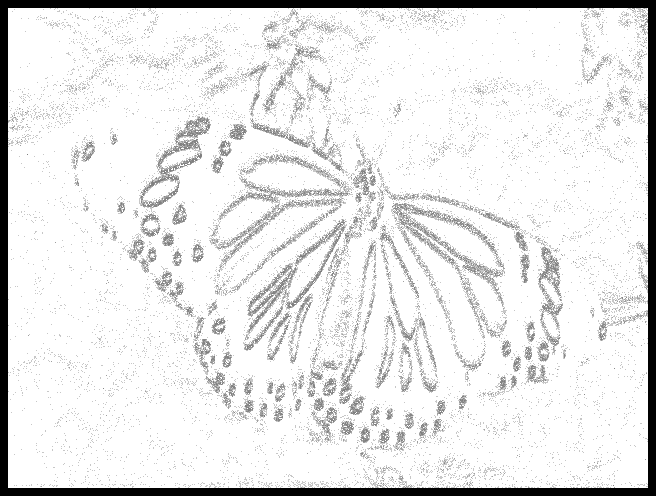}
            \caption{Variant 7}
        \end{subfigure}\hspace{-2mm}
        \begin{subfigure}{0.315\linewidth}
            \includegraphics[width=\linewidth]{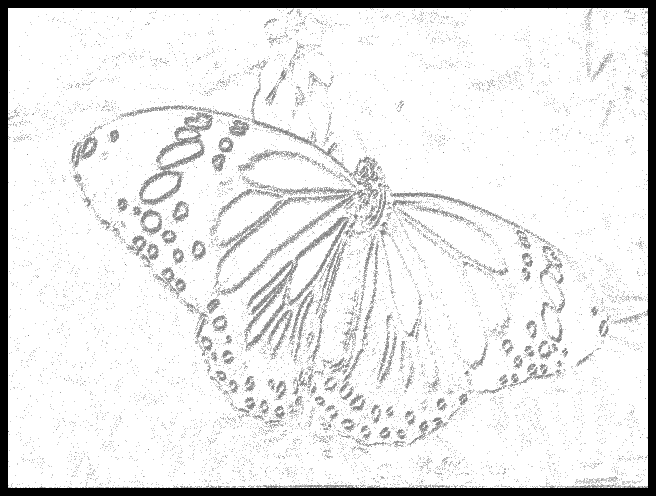}
            \caption{Variant 9}
        \end{subfigure}
    \end{minipage}
    \vspace{\figcapmargin}
    \caption{
    An N-ImageNet~\citep{DiST-N-IN} sample from the ``monarch butterfly" category.
    We show (a) the source RGB image from ImageNet~\citep{ImageNet} used to generate the events for reference.
    We plot (b) the event histograms of this data under the normal capture setting and (c-h) several robustness variants.
    Variant 2, 3, and 5 introduce changes in the camera trajectory, while Variant 6, 7, and 9 use different lighting during data capture.
    \textit{Left}: we apply the red-blue color map for better visual quality.
    \textit{Right}: the actual inputs to \algoNameFull with the gray-scale color map.
    }
    \label{app-fig:event-vis-butterfly}
    \vspace{\figmargin}
\end{figure*}

In \cref{app-fig:event-vis-butterfly}, we visualize data from the N-ImageNet test set and its robustness variants.
The original event image presents the best visual quality with both sharp object boundaries and small background noises.
This is because the original data are captured under a regular camera trajectory (square) and a small moving displacement.
In contrast, with an irregular moving trajectory (Variant 2, only horizontal movement), some boundaries are missing due to small image gradients along the moving direction.
With a larger moving amplitude (Variant 3, 5), the event images show severe motion blur, and there are lots of background noisy events.
For lighting changes, both too-low (Variant 6, 7) and too-high (Variant 9) illuminations result in distorted object boundaries.
Overall, these data variations cause significant train-test discrepancy, leading to a large performance drop in event-based classifiers trained solely on event camera datasets.
On the contrary, CLIP is trained on Internet-scale data covering diverse environmental conditions, which greatly improves the robustness of \algoNameFull.
Indeed, we successfully classify all variants of this data, while DiST fails on Variant 5, 6, and 7.

\begin{figure*}[t]
    \vspace{-8mm}
    \begin{minipage}{.5\linewidth}
        \centering
        \begin{subfigure}{0.47\linewidth}
            \includegraphics[width=\linewidth]{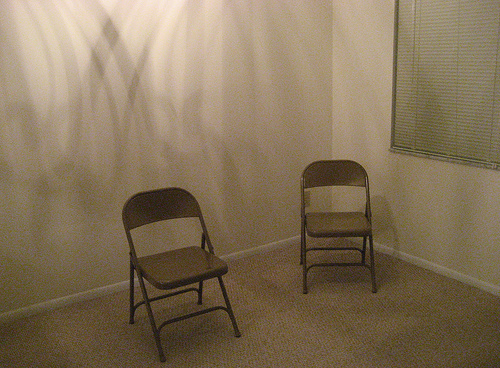}
            \caption{RGB image}
        \end{subfigure}\hspace{-2mm}
        \begin{subfigure}{0.47\linewidth}
            \includegraphics[width=\linewidth]{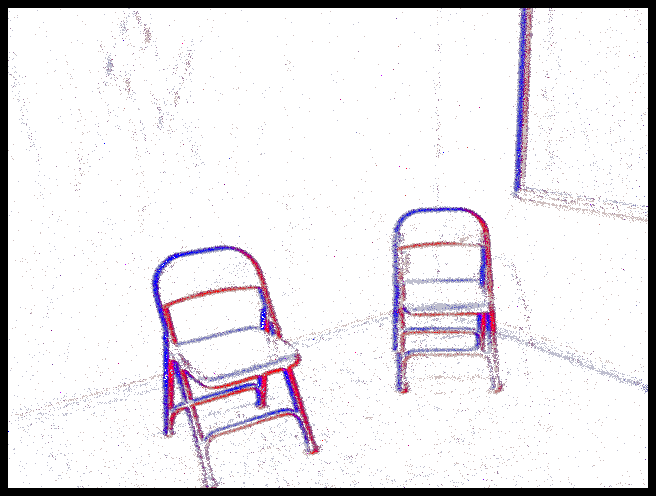}
            \caption{Original}
        \end{subfigure}
        \\
        \begin{subfigure}{0.315\linewidth}
            \includegraphics[width=\linewidth]{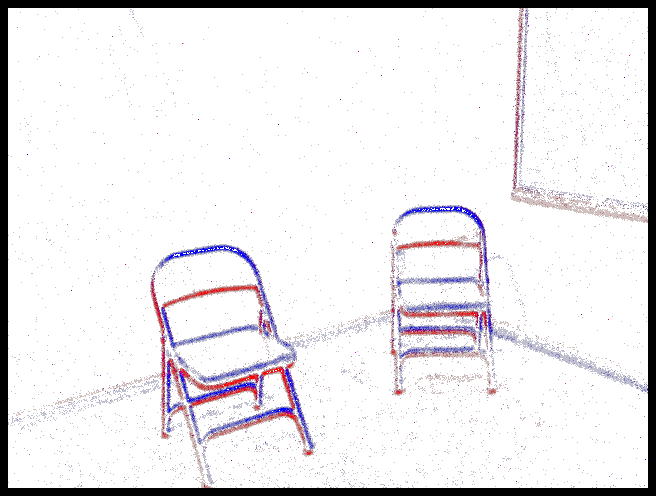}
            \caption{Variant 2}
        \end{subfigure}\hspace{-2mm}
        \begin{subfigure}{0.315\linewidth}
            \includegraphics[width=\linewidth]{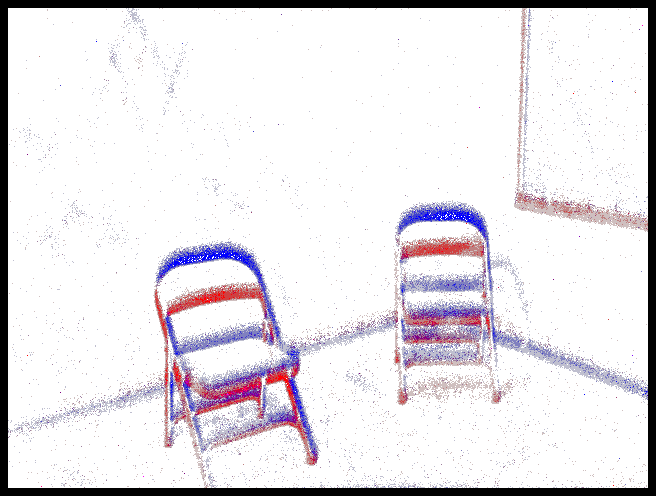}
            \caption{Variant 2}
        \end{subfigure}\hspace{-2mm}
        \begin{subfigure}{0.315\linewidth}
            \includegraphics[width=\linewidth]{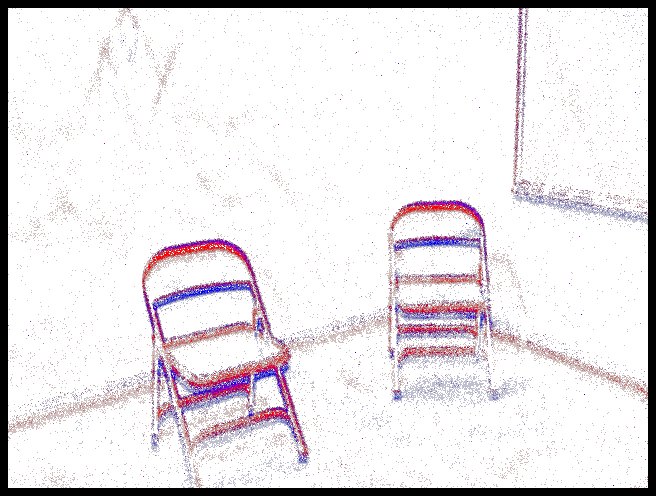}
            \caption{Variant 5}
        \end{subfigure}
        \\
        \begin{subfigure}{0.315\linewidth}
            \includegraphics[width=\linewidth]{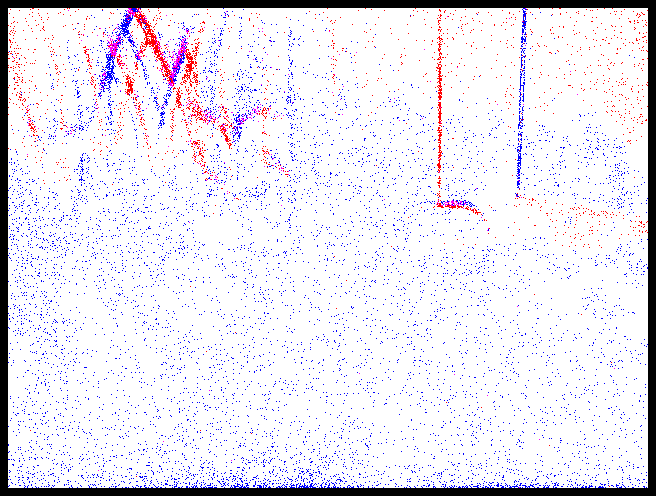}
            \caption{Variant 6}
        \end{subfigure}\hspace{-2mm}
        \begin{subfigure}{0.315\linewidth}
            \includegraphics[width=\linewidth]{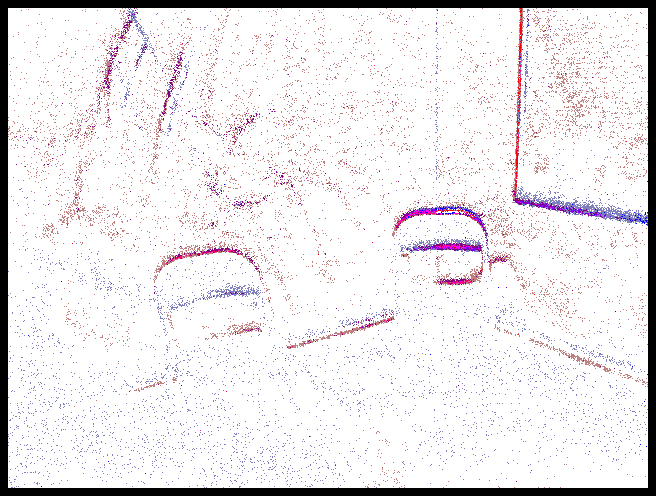}
            \caption{Variant 7}
        \end{subfigure}\hspace{-2mm}
        \begin{subfigure}{0.315\linewidth}
            \includegraphics[width=\linewidth]{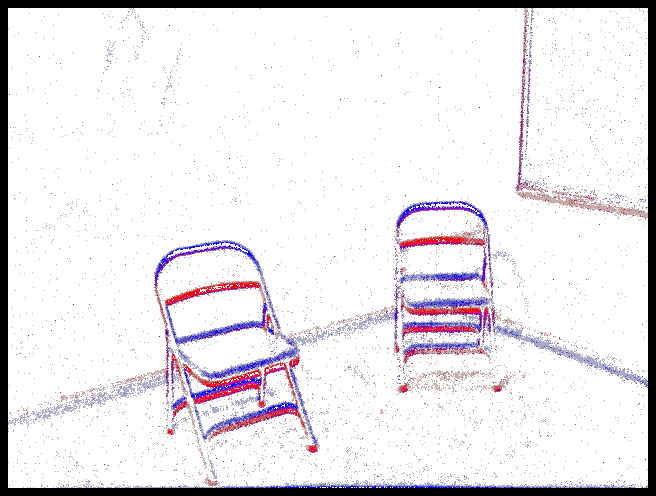}
            \caption{Variant 9}
        \end{subfigure}
    \end{minipage}
    \begin{minipage}{.5\linewidth}
        \centering
        \begin{subfigure}{0.47\linewidth}
            \includegraphics[width=\linewidth]{imgs/qual/folding-chair_27953/ILSVRC2012_val_00002253.JPEG}
            \caption{RGB image}
        \end{subfigure}\hspace{-2mm}
        \begin{subfigure}{0.47\linewidth}
            \includegraphics[width=\linewidth]{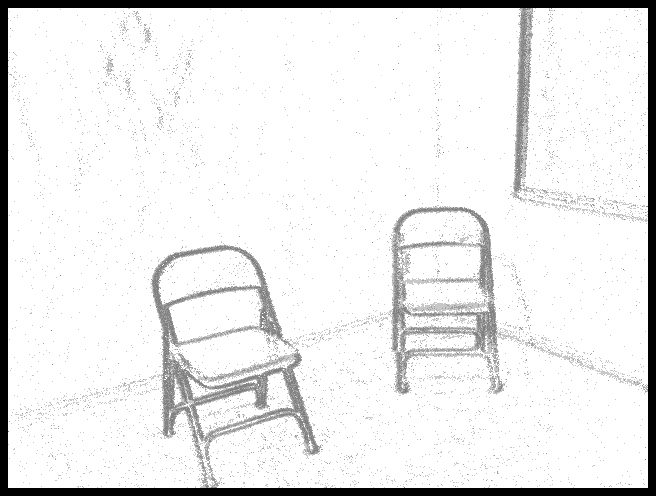}
            \caption{Original}
        \end{subfigure}
        \\
        \begin{subfigure}{0.315\linewidth}
            \includegraphics[width=\linewidth]{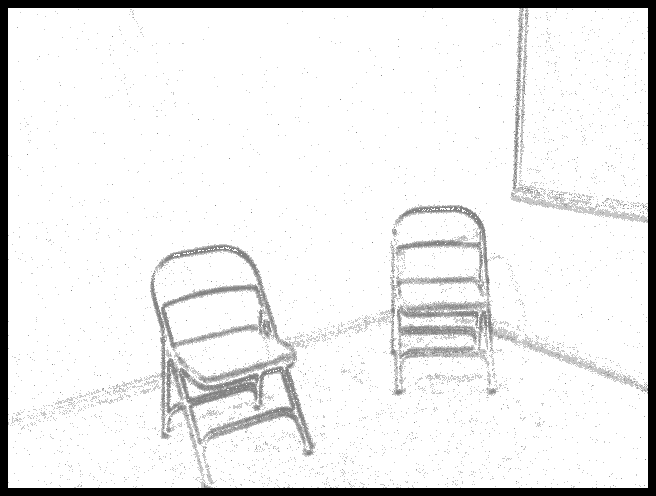}
            \caption{Variant 2}
        \end{subfigure}\hspace{-2mm}
        \begin{subfigure}{0.315\linewidth}
            \includegraphics[width=\linewidth]{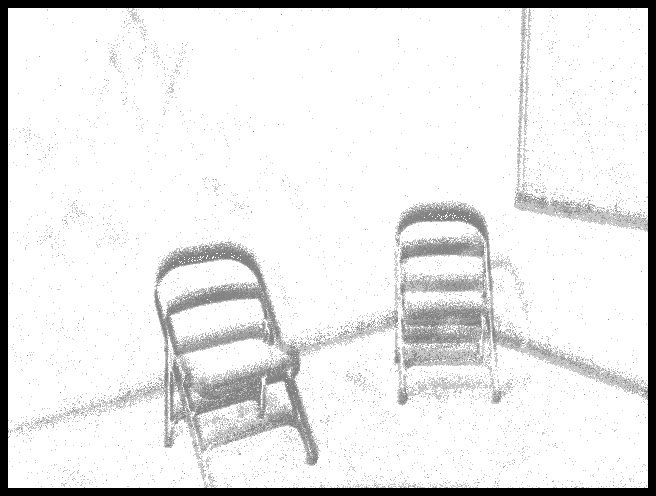}
            \caption{Variant 2}
        \end{subfigure}\hspace{-2mm}
        \begin{subfigure}{0.315\linewidth}
            \includegraphics[width=\linewidth]{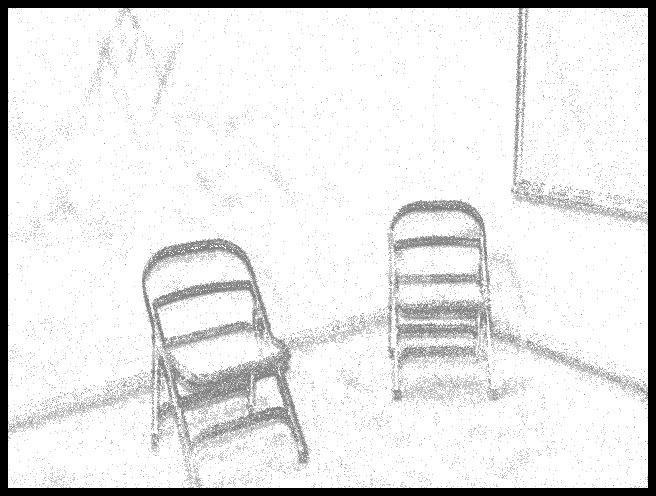}
            \caption{Variant 5}
        \end{subfigure}
        \\
        \begin{subfigure}{0.315\linewidth}
            \includegraphics[width=\linewidth]{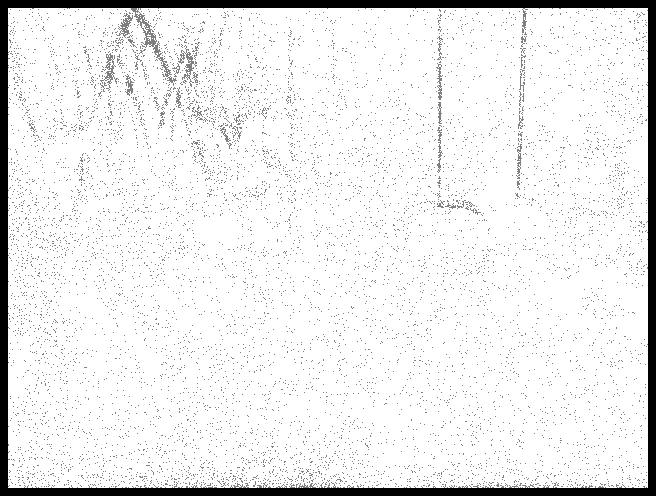}
            \caption{Variant 6}
        \end{subfigure}\hspace{-2mm}
        \begin{subfigure}{0.315\linewidth}
            \includegraphics[width=\linewidth]{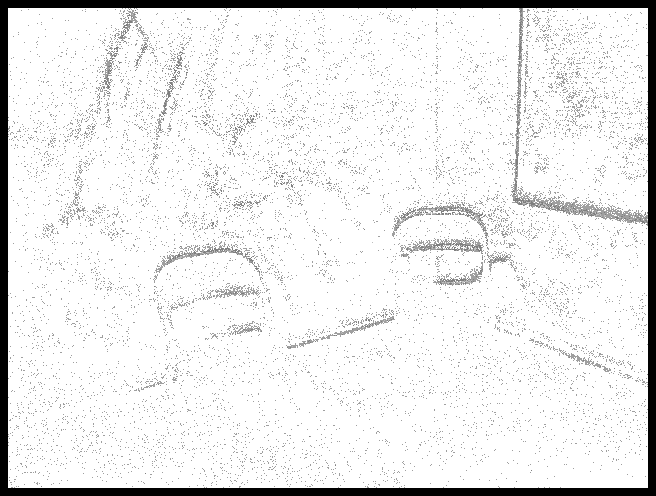}
            \caption{Variant 7}
        \end{subfigure}\hspace{-2mm}
        \begin{subfigure}{0.315\linewidth}
            \includegraphics[width=\linewidth]{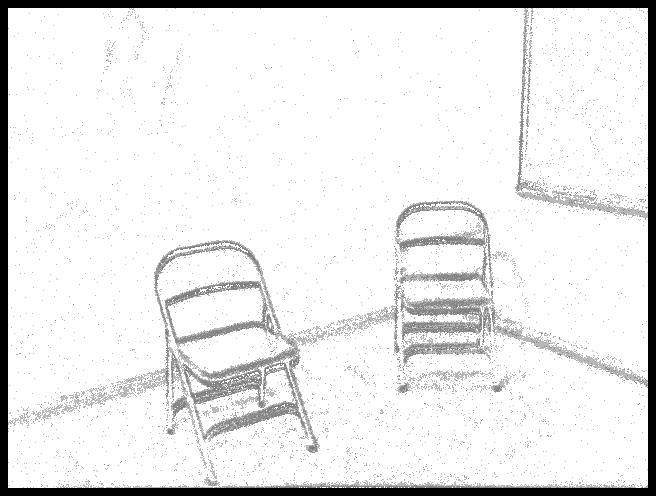}
            \caption{Variant 9}
        \end{subfigure}
    \end{minipage}
    \vspace{\figcapmargin}
    \caption{
    An N-ImageNet sample from the ``folding chair" category.
    We again plot the red-blue colorized images at left and the gray-scale input to \algoNameFull at right.
    Due to extreme low light, events under Variant 6 do not contain any object information, making classification impossible.
    }
    \label{app-fig:event-vis-chair}
    \vspace{-2mm}
\end{figure*}

We show another example in \cref{app-fig:event-vis-chair}.
The event images under different camera motions follow similar distortions.
However, under the low-light condition (Variant 6, 7), the chairs almost disappear, making the recognition task problematic.
As a result, neither \algoNameFull nor baselines is able to predict the correct category.

\begin{table}[t]
    \vspace{1mm}
    \centering
    \small
    \setlength{\tabcolsep}{3pt}
    \begin{tabular}{ccccccc}
        \toprule
        Labeled Data per Class & 1 & 3 & 5 & 10 & 20 \\
        \midrule
        Few-shot only & 30.54 & 32.64 & 32.92 & 37.04 & 39.58 \\
        Few-shot + Pseudo-labels & \textbf{35.98} & \textbf{37.04} & \textbf{37.53} & \textbf{40.30} & \textbf{42.02} \\
        \bottomrule
    \end{tabular}
    \vspace{\tablecapmargin}
    \caption{
    \textbf{Semi-supervised learning results}.
    We report the accuracy of \algoNameFull trained only on few-shot labeled data \vs few-shot labeled data plus generated pseudo-labels.
    \label{app-table:semi-sup-eventclip}
    }
    \vspace{-2mm}
\end{table}

\subsection{Semi-Supervised Learning with \algoNameFull}\label{app:semi-sup-eventclip}

We conduct semi-supervised learning on N-ImageNet (Mini), where a few labeled data are available per class, and we pseudo-label the remaining unlabeled data.
As shown in \cref{app-table:semi-sup-eventclip}, training on generated pseudo-labels consistently improves the accuracy compared to using only labeled data, showing a promising direction of leveraging raw data.

\begin{table}[t]
    \vspace{1mm}
    \centering
    \small
    \setlength{\tabcolsep}{3pt}
    \begin{tabular}{ccc}
        \toprule
        & Event-to-Frame (CPU) & Model Forward (GPU) \\
        \midrule
        Time (ms) & 6.76 & 8.73 \\
        \bottomrule
    \end{tabular}
    \vspace{\tablecapmargin}
    \caption{
    \textbf{Speed analysis of \algoNameFull on N-ImageNet}.
    The event camera resolution is 640$\times$480.
    We report the runtime (ms) of \algoNameFull's two main components.
    \label{app-table:eventclip-speed}
    }
    \vspace{-3mm}
\end{table}

\subsection{Speed Analysis of \algoNameFull}\label{app:model-speed}

Low imaging latency is an important property of event cameras.
We test the inference time of our \algoNameFull pipeline, which consists of two components: event-to-frame conversion (CPU) and model forward pass (GPU).
All speeds are measured on a Linux desktop with AMD Ryzen 9 5950X CPU (16-Core) and NVIDIA GeForce 3090 GPU.
We report the runtime on N-ImageNet which has the highest frame resolution of 640$\times$480 in \cref{app-table:eventclip-speed}.
Due to the use of simple event counting, our event-to-frame conversion time is 6.76 ms.
Adding to the 8.73 ms model forward time, \algoNameFull is able to run at more than 50 FPS, rendering it suitable for real-time applications with event cameras.

\begin{figure}[t]
    \centering
    \begin{subfigure}{0.95\linewidth}
        \includegraphics[width=\linewidth]{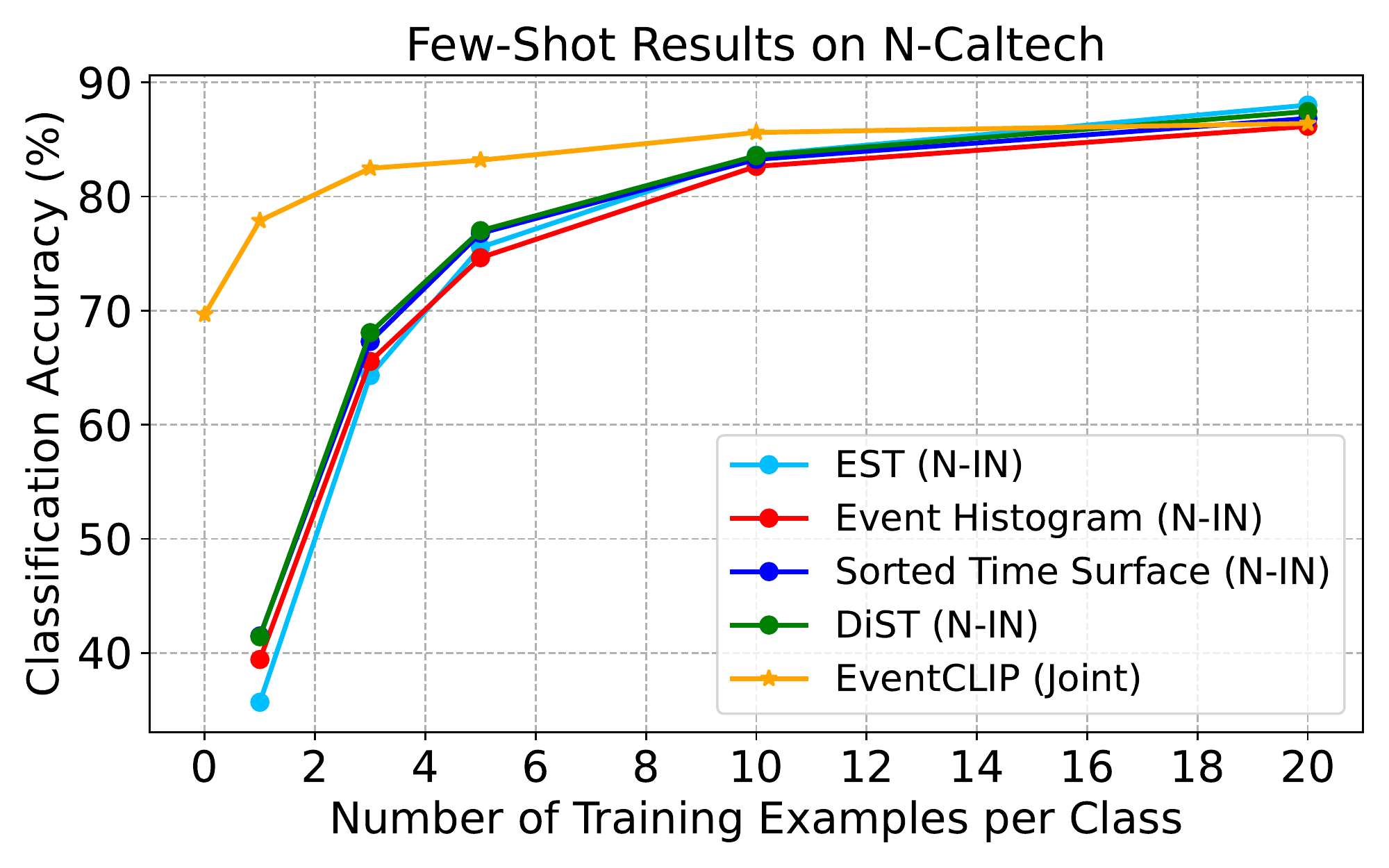}
    \end{subfigure}
    \\
    \begin{subfigure}{0.95\linewidth}
        \includegraphics[width=\linewidth]{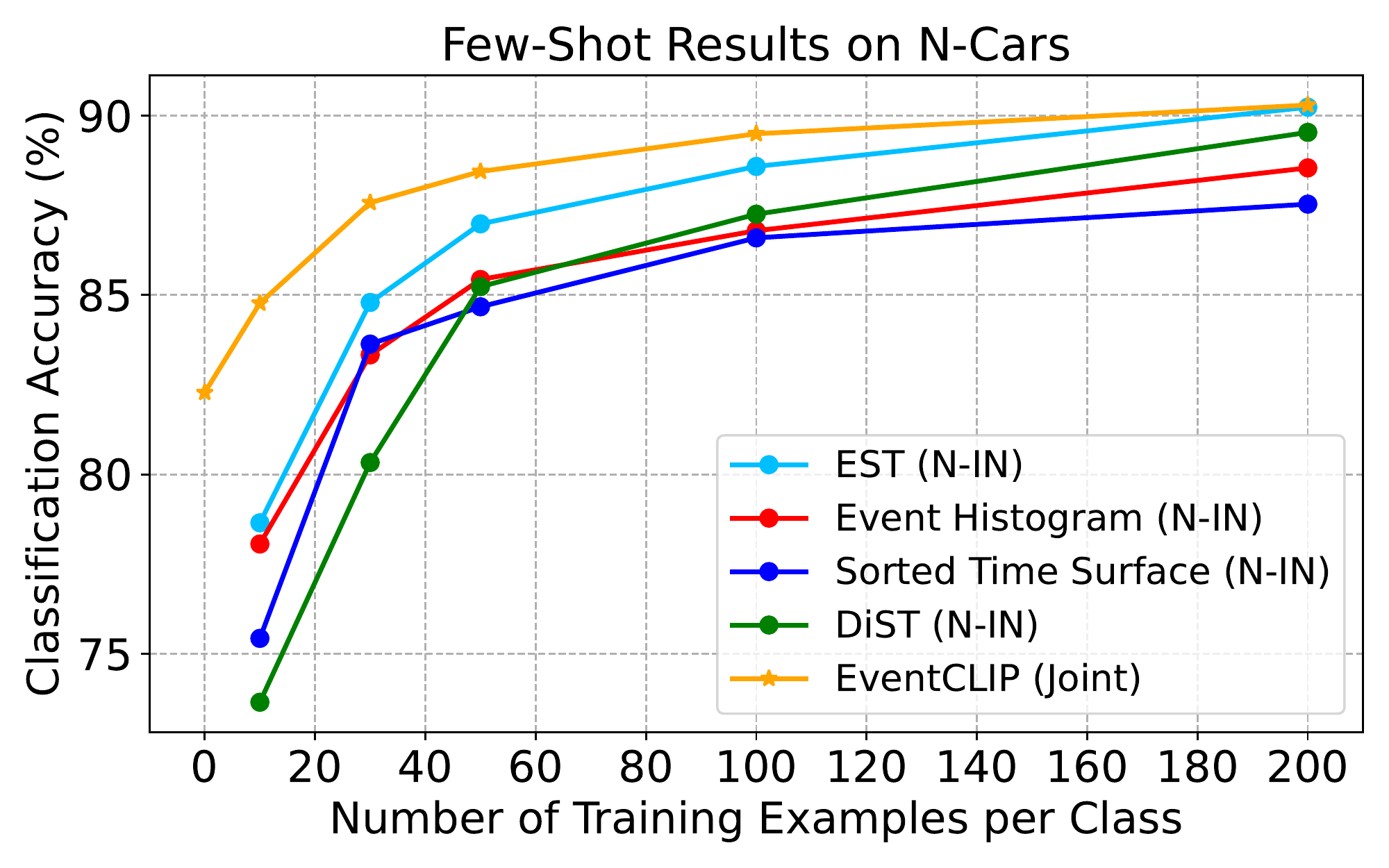}
    \end{subfigure}
    \vspace{\figcapmargin}
    \caption{
    Comparison of few-shot accuracy between \algoNameFull and baselines initialized from weights pre-trained on N-ImageNet.
    Our method still achieves more data-efficient learning on N-Cars, while performing competitively on N-Caltech.
    }
    \label{app-fig:few-shot-n-in}
    \vspace{\figmargin}
\end{figure}

\subsection{N-ImageNet Pre-trained Baselines}\label{app:nin-pretrain-baseline}

In our main experiments, all baselines employ backbones pre-trained on the RGB images from ImageNet.
Here, we evaluate a more challenging scenario, where the backbones are \textit{initialized from weights pre-trained on the large-scale N-ImageNet dataset}.
\cref{app-fig:few-shot-n-in} compares the few-shot accuracy of \algoNameFull and the baselines.
N-ImageNet pre-training provides substantial domain-specific knowledge for event-based classification, and thus greatly improves their performance.
Still, \algoNameFull is able to outperform the baselines with a sizeable margin when the number of data per category is small (\eg 10-shot), and achieve competitive results with more training data.
This demonstrates the effective knowledge transfer process of our approach.

\clearpage

\clearpage

\begin{figure*}[t]
    \vspace{\pagetopmargin}
    \begin{minipage}{.27\linewidth}
        \centering
        \begin{subfigure}{\linewidth}
            \includegraphics[width=\linewidth]{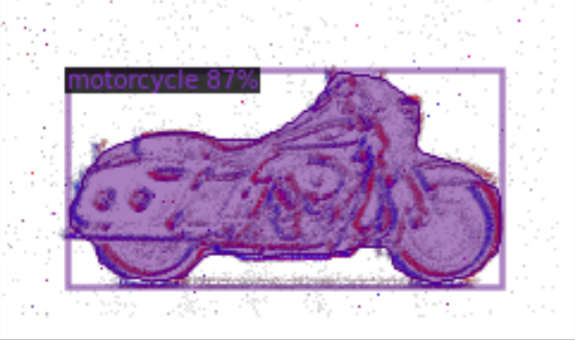}
            \caption{N-ImageNet sample}
        \end{subfigure}\\
        \begin{subfigure}{\linewidth}
            \includegraphics[width=\linewidth]{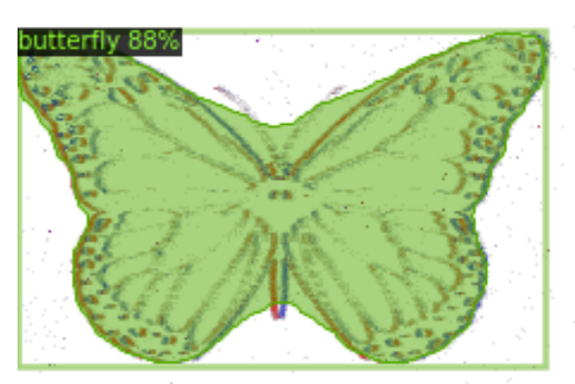}
            \caption{N-ImageNet sample}
        \end{subfigure}
    \end{minipage}
    \hfill
    \begin{minipage}{.71\linewidth}
        \centering
        \begin{subfigure}{\linewidth}
            \includegraphics[width=\linewidth]{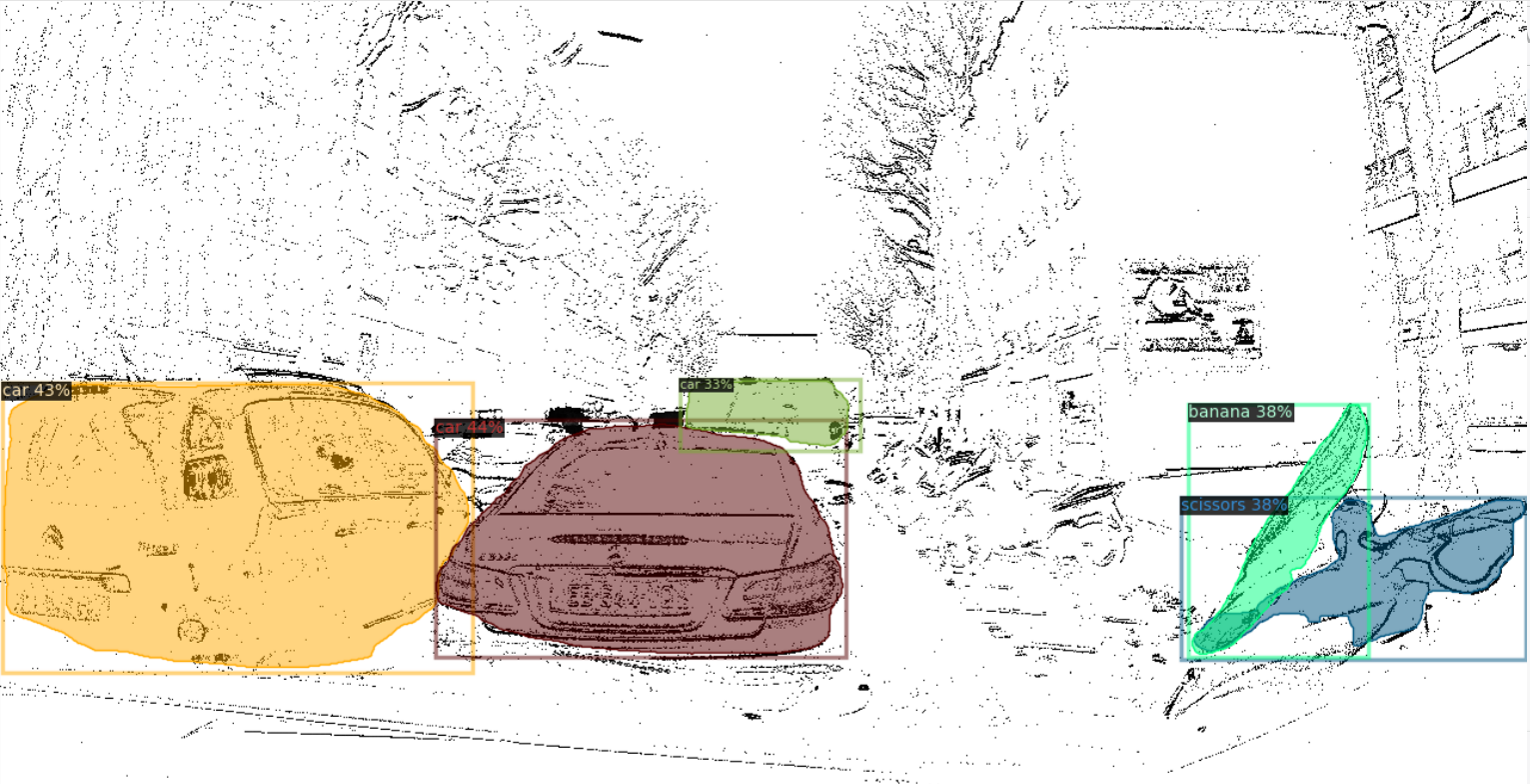}
            \caption{Gen1~\citep{Gen1Det} detection data}
        \end{subfigure}
    \end{minipage}
    \vspace{\figcapmargin}
    \caption{
    Transfer Detic~\citep{Detic} to perform instance segmentation on (a), (b) synthetic data, and (c) real-world detection data.
    }
    \label{app-fig:det-seg}
    \vspace{\figmargin}
\end{figure*}

\begin{table}[h]
    \centering
    \small
    \setlength{\tabcolsep}{4pt}
    \begin{tabular}{ccccccc}
        \toprule
        \textbf{Dataset} & \multicolumn{6}{c}{N-Caltech} \\
        Shots & 0 & 1 & 3 & 5 & 10 & 20 \\
        \midrule
        Visual-MLP & 69.67 & 73.52 & 77.43 & 78.55 & 82.69 & 83.28 \\
        Visual-Trans. & 69.67 & 76.51 & 81.44 & 82.99 & 85.21 & 85.73 \\
        Text & 69.67 & 77.35 & 81.58 & 82.81 & 84.89 & 86.33 \\
        Joint & 69.67 & 77.89 & 82.48 & 83.19 & 85.62 & 86.41 \\
        \bottomrule
    \end{tabular}
    \vspace{\tablecapmargin}
    \caption{
    Full zero-shot and few-shot classification accuracy ($\%$) of \algoNameFull on N-Caltech.
    \label{app-table:ncaltech-full}
    }
    \vspace{\tablemargin}
    \vspace{-1mm}
\end{table}

\begin{table}[h]
    \centering
    \small
    \setlength{\tabcolsep}{4pt}
    \begin{tabular}{ccccccc}
        \toprule
        \textbf{Dataset} & \multicolumn{6}{c}{N-Cars} \\
        Shots & 0 & 10 & 30 & 50 & 100 & 200 \\
        \midrule
        Visual-Trans. & 82.28 & 84.55 & 86.62 & 87.36 & 89.51 & 90.33 \\
        Text & 82.28 & 84.37 & 87.06 & 87.20 & 89.40 & 90.34 \\
        Joint & 82.28 & 84.77 & 87.57 & 88.44 & 89.49 & 90.29 \\
        \bottomrule
    \end{tabular}
    \vspace{\tablecapmargin}
    \caption{
    Full zero-shot and few-shot classification accuracy ($\%$) of \algoNameFull on N-Cars.
    \label{app-table:ncars-full}
    }
    \vspace{\tablemargin}
    \vspace{-1mm}
\end{table}

\begin{table}[h]
    \centering
    \small
    \setlength{\tabcolsep}{4pt}
    \begin{tabular}{ccccccc}
        \toprule
        \textbf{Dataset} & \multicolumn{6}{c}{N-ImageNet} \\
        Shots & 0 & 1 & 3 & 5 & 10 & 20 \\
        \midrule
        Visual-Trans. & 20.78 & 23.36 & 24.45 & 25.20 & 26.64 & 28.45 \\
        Text & 20.78 & 24.04 & 25.47 & 26.55 & 28.24 & 30.25 \\
        Joint & 20.78 & 24.25 & 25.84 & 26.96 & 28.63 & 30.57 \\
        \bottomrule
    \end{tabular}
    \vspace{\tablecapmargin}
    \caption{
    Full zero-shot and few-shot classification accuracy ($\%$) of \algoNameFull on N-ImageNet.
    \label{app-table:nin-full}
    }
    \vspace{\tablemargin}
\end{table}

\subsection{Full Numerical Results}\label{app:full-result-tables}

In the main paper, we plot \algoNameFull's zero-shot and few-shot classification accuracy in Fig.~{\color{red} 3} and Fig.~{\color{red} 4}.
To ease future comparison, we report all those numbers in \cref{app-table:ncaltech-full}, \cref{app-table:ncars-full}, and \cref{app-table:nin-full}.

\section{Limitations and Future Works}\label{app:limitations}

In this paper, we mainly focus on the event-based object recognition problem.
It is still unclear how to utilize large pre-trained models for other event camera tasks, such as detection~\citep{Gen1Det,1MpxDet} and segmentation~\citep{Evdistill,ESS_EvSegFromImg}.
We conduct preliminary experiments to transfer an open-world instance segmentation model Detic~\citep{Detic} to event data.
Since Detic also builds upon CLIP, it is able to detect objects from N-ImageNet samples as shown in \cref{app-fig:det-seg} (a), (b).
However, objects present more complex motions in real-world captured events, degrading the converted event frames' visual quality drastically.
Still, Detic is able to detect some objects as shown in \cref{app-fig:det-seg} (c).
But it also misses objects with sparse edges such as the motorbikes and trucks.
Besides, it predicts some weird classes such as bananas and scissors due to the large domain gap.
Therefore, it is worth studying better event-to-frame conversion or model adaptation methods to better leverage the pre-trained vision foundation models.

Another direction is to directly learn a joint embedding space of events and texts.
Due to a lack of event-text dataset, we can leverage RGB images as the intermediate representation, as done in some recent work~\citep{ImageBind,ULIP}.
We can first leverage the large-scale RGB video datasets such as CO3D~\citep{CO3D} to simulate event data.
Then, we train an event encoder to align the extracted features with a pre-trained image encoder such as CLIP.

\end{document}